\documentclass[lettersize,journal,compsoc]{IEEEtran}
\usepackage{amsmath,amsfonts}
\usepackage{algorithmic}
\usepackage{algorithm}
\usepackage{array}
\usepackage[caption=false,font=normalsize,labelfont=sf,textfont=sf]{subfig}
\usepackage{textcomp}
\usepackage{stfloats}
\usepackage{url}
\usepackage{verbatim}
\usepackage{graphicx}
\usepackage{cite}
\usepackage{booktabs}       % professional-qualit\y tables
\usepackage{nicefrac}       % compact symbols for 1/2, etc.
\usepackage{microtype}      % microtypography
\usepackage[table]{xcolor}
\usepackage[misc]{ifsym} 
\usepackage{hyperref}

\usepackage{amssymb}
\usepackage{color}
\usepackage{multirow}
\usepackage{bbding}
\usepackage{longtable}
\usepackage{supertabular}

\begin{document}

\title{Appeal: Allow Mislabeled Samples the Chance to be Rectified in Partial Label Learning}

\author{Chongjie Si, Xuehui Wang, Yan Wang, Xiaokang Yang,~\IEEEmembership{~Fellow,~IEEE},  Wei Shen
        % <-this % stops a space
\IEEEcompsocitemizethanks{
\IEEEcompsocthanksitem This work was supported by NSFC 62176159, 62322604, Natural Science Foundation of Shanghai 21ZR1432200, and Shanghai Municipal Science and Technology Major Project 2021SHZDZX0102.
\IEEEcompsocthanksitem C. Si, X. Wang, X. Yang, W. Shen are with MoE Key Lab of Artificial Intelligence, AI Institute, Shanghai Jiao Tong University, Shanghai 200240, China. \protect E-mail: \{chongjiesi, wangxuehui, xkyang, wei.shen\}@sjtu.edu.cn.
\IEEEcompsocthanksitem Y. Wang is with the Shanghai Key Lab of Multidimensional Information Processing, East China Normal University, Shanghai 200241, China. \protect E-mail: ywang@cee.ecnu.edu.cn.
}
}

% The paper headers
\markboth{Journal of \LaTeX\ Class Files,~Vol.~14, No.~8, August~2021}%
{Shell \MakeLowercase{\textit{et al.}}: A Sample Article Using IEEEtran.cls for IEEE Journals}

%\IEEEpubid{0000--0000/00\$00.00~\copyright~2021 IEEE}
% Remember, if you use this you must call \IEEEpubidadjcol in the second
% column for its text to clear the IEEEpubid mark.

\IEEEtitleabstractindextext{
\begin{abstract}
In partial label learning (PLL), each instance is associated with a set of candidate labels among which only one is ground-truth. The majority of the existing works focuses on constructing robust classifiers to estimate the labeling confidence of candidate labels in order to identify the correct one. However, these methods usually struggle to identify and rectify mislabeled samples. To help these mislabeled samples ``appeal'' for themselves and help existing PLL methods identify and rectify mislabeled samples, in this paper, we propose the first appeal-based PLL framework. Specifically, we introduce a novel partner classifier and instantiate it predicated on the implicit fact that non-candidate labels of a sample should not be assigned to it, which is inherently accurate and has not been fully investigated in PLL. Furthermore, a novel collaborative term is formulated to link the base classifier and the partner one. During each stage of mutual supervision, both classifiers will blur each other's predictions through a blurring mechanism to prevent overconfidence in a specific label. Extensive experiments demonstrate that the appeal and disambiguation ability of several well-established stand-alone and deep-learning based PLL approaches can be significantly improved by coupling with this learning paradigm.

\end{abstract}

}

\maketitle
\IEEEdisplaynontitleabstractindextext
\IEEEpeerreviewmaketitle

\section{Introduction}

\begin{figure}
    \centering
    \includegraphics[scale=0.15]{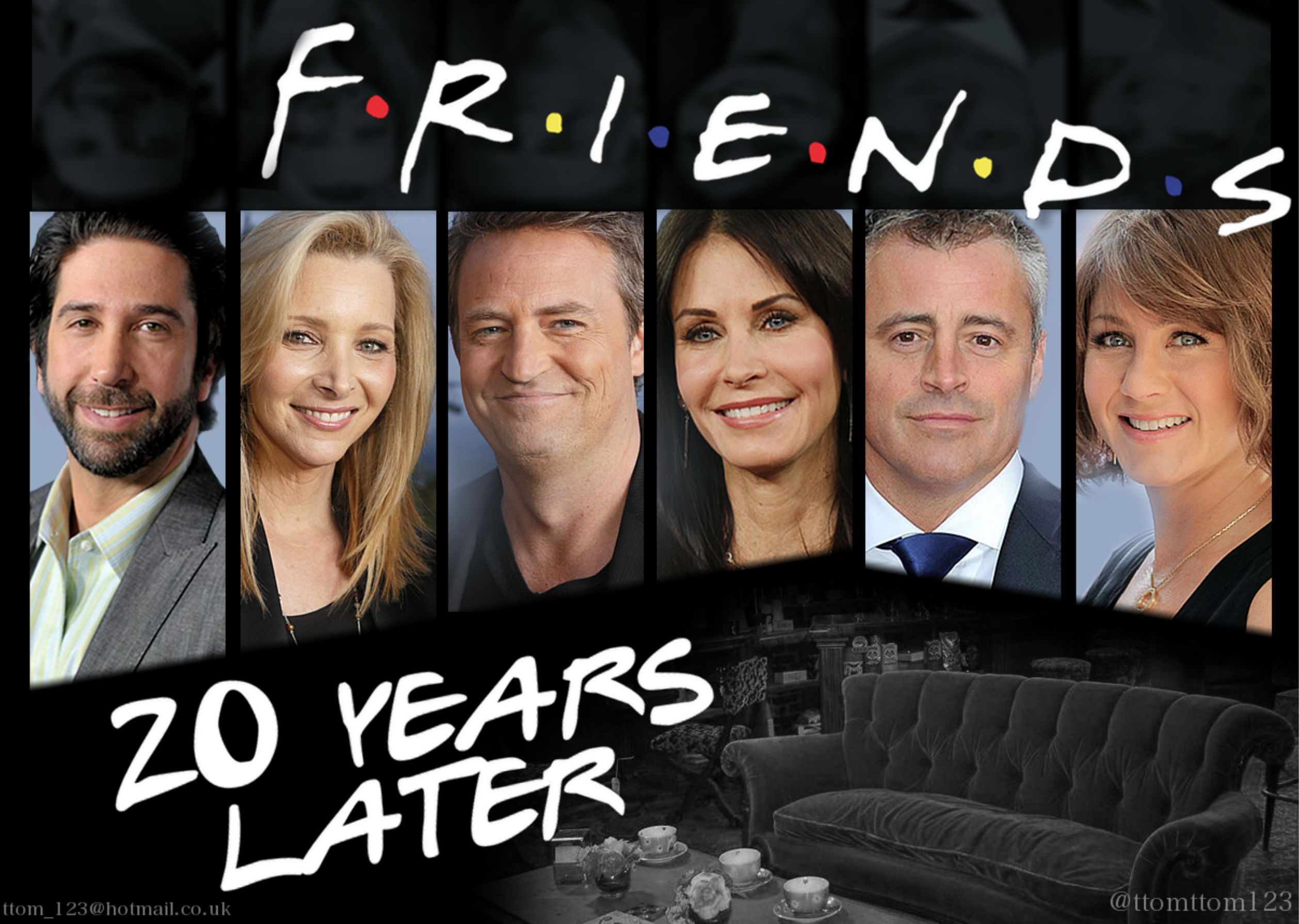}
    \caption{Partial label learning in the automatic face naming task. The faces in this image can be automatically detected by a face detector, and then each face is assigned with the candidate name label set extracted from the script: Jennifer Aniston, Courteney Cox, Lisa Kudrow, Matthew Perry, David Schwimmer, and Matt LeBlanc, where only one is ground-truth.}
    \label{fig:example PLL}
\end{figure}

In 2002, Jin and Ghahramani \cite{jin2002learning} pioneered the concept of Partial Label Learning (PLL), a groundbreaking framework in which each instance is associated with multiple candidate labels, among which only one represents the actual ground-truth. Over the past two decades, the field of PLL has experienced remarkable growth \cite{cour2011learning,han2018co22,jin2002learning,papandreou2015weaklysfs,ren2018learning,zhou2018brief,zhu2009introductionddd,chai2019large2462, li2019towards, li2019safe}, driven by the increasing necessity to accurately discern the valid label from a set of potential candidates in diverse real-world applications. A notable example, as shown in Fig. \ref{fig:example PLL}, 
is the automatic face naming task \cite{zeng2013learningSoccerplayer,guillaumin2010multipleYahoonews}, where each facial image extracted from various media is linked to a list of names derived from corresponding titles or captions \cite{gong2022partialicml}. Another salient application is facial age estimation: for each human face, the ages annotated by crowd-sourcing labelers are considered as candidate labels \cite{panis2016overviewFg-net}.

\begin{figure*}
  \centering
  \subfloat[]{
		\includegraphics[scale=0.17]{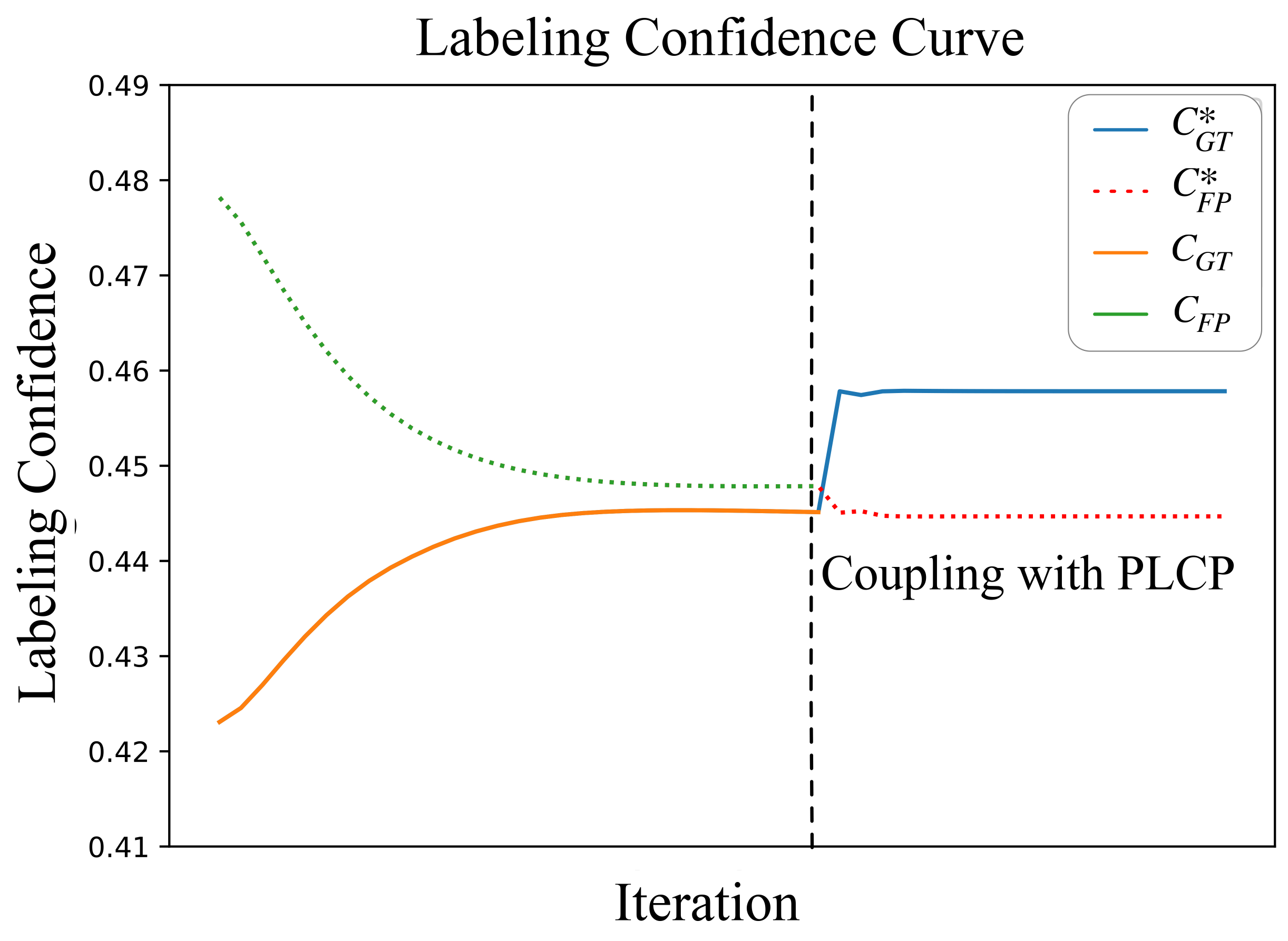}}
    \subfloat[]{
		\includegraphics[scale=0.17]{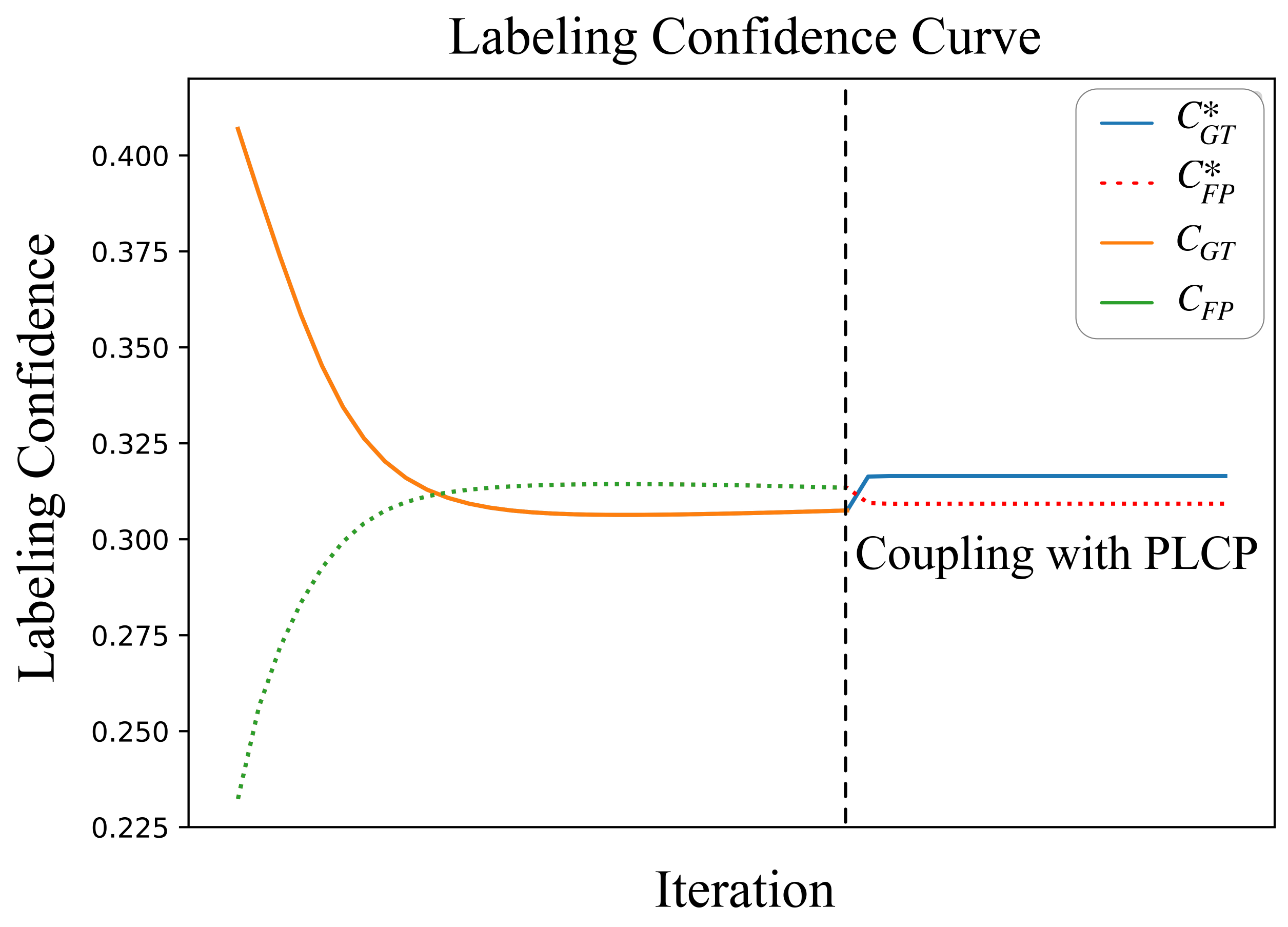}}
  \caption{Two representative errors a typical PLL classifier, e.g., PL-AGGD \cite{wang2021adaptivePLAGGD} may make. $C_{GT}$ and $C_{GT}^*$ stand for the labeling confidence of the ground-truth label generated by PL-AGGD and PL-AGGD coupled with PLCP, and $C_{FP}$ and $C_{FP}^*$ stand for that of a false positive label predicted by PL-AGGD and PL-AGGD coupled with PLCP. (a). For a false positive candidate label with a large labeling confidence, although its confidence may decrease properly, it could still be larger than the ground-truth one's. (b). The labeling confidence of a false positive candidate label keeps increasing and becomes the largest, which misleads the final prediction. When coupled with PLCP, the labeling confidence of each candidate label generated by the partner classifier is adopted as the supervision to help PL-AGGD correct these errors, which results in a mutation in the figures.}
  \label{fig:false positive}
\end{figure*}

Evidently, the principal challenge in PLL lies in the inherent obscurity of the ground-truth label, which conceals within the candidate labels and remains inaccessible during the training phase. 
To surmount this obstacle, the most extensively researched and pivotal strategy in PLL-\textit{\textbf{Disambiguation}}-emerges as critical. Existing methods predominantly achieve disambiguation by differentiating the labeling confidences of each candidate label to identify the ground truth. This process typically relies on an alternative and iterative optimization algorithm for updating the classifier's parameters. For instance, LALO \cite{2018Leveraginglalo} implements constrained local consistency to differentiate the candidate labels, PL-AGGD \cite{wang2021adaptivePLAGGD} employs a similarity graph for effective disambiguation and PL-CL \cite{jia2023complementary} adopts complementary information to help disambiguation. However, a significant yet rarely studied question comes to the front: \textit{can the instances that are incorrectly disambiguated, i.e., mislabeled, have the opportunity to be rectified?} More specifically, \textit{can a classifier correct a false positive candidate label (i.e., invalid candidate label) with a large or upward-trending labeling confidence at a later stage?} To explore this question, we conduct experiments on a real-world data set Lost \cite{cour2009learningLost} and record the labeling confidences of several candidate labels generated by PL-AGGD \cite{wang2021adaptivePLAGGD} in each iteration, which is shown in Fig. \ref{fig:false positive}. Our findings reveal some intriguing phenomena: 
\begin{itemize}
    \item Each candidate label's labeling confidence is likely to continually increase or decrease until convergence.
    \item For a false positive candidate label with a large labeling confidence, although its confidence may decrease in subsequent iterations, the confidence remains substantial and can easily lead to the incorrect identification of the ground truth label.
\end{itemize}

The observed phenomena suggest that once the labeling confidence of a false positive candidate label increases, it becomes difficult to decrease in the subsequent iterations. Furthermore, even if the confidence of a false positive candidate label decreases appropriately, it may still be recognized as the ground truth one, as its initial labeling confidence remains large and continues to be greater than the confidence of the ground truth label upon convergence. As a result, correcting mislabeled samples for a PLL classifier itself proves to be quite challenging.

We believe that these mislabeled samples should be afforded the opportunity to be corrected. In other words, these samples should have a chance to ``appeal'' on their behalf. In this case, the classifier's disambiguation ability can also be further enhanced. Consequently, in this paper, we introduce another major strategy in PLL: \textit{\textbf{Appeal}}, and propose the first appeal-based framework.

Specifically, our framework PLCP, referring to \textbf{P}artial \textbf{L}abel Learning with a \textbf{C}lassifier as  \textbf{P}artner, provides each instance with the chance to appeal. Given a classifier from any existed PLL approach as the base classifier, as shown in Fig. \ref{fig:framework_plcp}, PLCP integrates an additional partner classifier. This partner classifier assists the base classifier in identifying and rectifying mislabeled samples, offering more precise and complementary information to the base classifier, thereby enhancing disambiguation and fostering mutual supervision between the two classifiers. The design of the partner classifier is crucial in whether mislabeled samples could appeal for themselves, as the partner classifier's feedback significantly influences the base classifier's capability to recognize and amend mislabeled instances. Since the information in non-candidate labels, which indicates that a set of labels DO NOT belong to a sample, is typically more precise yet often overlooked by the majority of existing works, the partner classifier is devised as a complementary classifier. This classifier specifies the labels that should not be assigned to a sample, thereby complementing the base classifier. Additionally, a collaborative term is also designed to link the base classifier and the partner classifier. 

During mutual supervision, the labeling confidence is first updated based on the base classifier's modeling output.  A blurring mechanism is then applied to this updated labeling confidence, introducing uncertainty. This could potentially diminish the high confidence of certain false positive candidate labels or elevate the low confidence of the actual ground truth. This updated labeling confidence subsequently serves as the supervision information to interact with the partner classifier, whose final output, in turn, supervises the base classifier. The predictions of the two classifiers, while distinct, are inextricably linked, enhancing the disambiguation ability of this paradigm in two opposing ways. With this mutual supervision paradigm, the instances with disambiguation errors have a higher likelihood to appeal successfully.

The preliminary results of this work are presented in \cite{si2023partial}. In this paper, we refine the method in \cite{si2023partial} and propose a more robust version of PLCP. Besides, comprehensive theoretical analyses of  PLCP are provided in this paper.
The rest of the paper is organized as follows. We first introduce the appeal-based framework PLCP in Sec. \ref{sec: method}, and then present the experimental results and ablation study in Sec. \ref{sec: exp}. Further analysis is provided in Sec. \ref{sec:fur}. We then review some related works in PLL and other related fields in Sec. \ref{sec:related}. Finally, conclusion is given in Sec. \ref{sec:con}
\section{Appeal-based Framework: PLCP}\label{sec: method}

Let $\mathcal{X}=\mathbb{R}^q$ denote the $q$-dimensional feature space and $\mathcal{Y} =\{0,1\}^l$ be the label space with $l$ classes. Given a partial label data set $\mathcal{D} = \{\mathbf{x}_i, S_i | 1\leq i \leq n\}$ where $\mathbf{x}_i \in \mathcal{X}$, $S_i \subseteq \mathcal{Y}$ is the corresponding candidate label set and $n$ is the number of instances. The task of PLL is to induce a multi-class classifier $f: \mathcal{X}\rightarrow \mathcal{Y}$ based on $\mathcal{D}$.

As shown in Fig. \ref{fig:framework_plcp}, a partner classifier $\mathcal{P}$ is designed to complement a base classifier $\mathcal{B}$ and also can be supervised by $\mathcal{B}$. $\mathcal{B}$ represents any existing PLL classifier with fine generality and flexibility. In each stage of mutual supervision, the base classifier updates the labeling confidence based on its modeling outputs, and then a blurring mechanism is applied to further process it. Subsequently the output of the labeling confidence is taken as the supervision information for the partner classifier. The learning pipeline of the partner classifier closely mirrors that of the base classifier. The following subsections will provide further details on this process.

\begin{figure*}[t] 
    \centering
    \includegraphics[scale=0.064]{pic/network.pdf}
    \caption{The framework of PLCP. A partner classifier is constructed based on the non-candidate label information to enable mutual supervision between the base classifier and itself. In each stage of mutual supervision, the base classifier updates the labeling confidence $\mathbf{P}$ based on its modeling output $\mathbf{M}$ and blurs it through a blurring mechanism. Afterwards, the output is represented as the supervision information to interact with the partner classifier. The pipeline of the partner classifier is almost the same as the base classifier's.}
    \label{fig:framework_plcp}
\end{figure*}

\subsection{Base Classifier}
Denote $\mathbf{X}=[\mathbf{x}_1,\mathbf{x}_2,...,\mathbf{x}_n]^\mathsf{T}\in\mathbb{R}^{n\times q}$ the sample matrix with $n$ instances, and $\mathbf{Y}=[\mathbf{y}_1, \mathbf{y_2},...,\mathbf{y}_n]^\mathsf{T}\in\{0,1\}^{n\times l}$ the partial label matrix, where $y_{ij}=1$ (resp. $y_{ij}=0$) if the $j$-th label of $\mathbf{x}_i$ resides in (resp. does not reside in) its candidate label set.
Suppose $\mathbf{P}=[\mathbf{p}_1, \mathbf{p}_2,...,\mathbf{p}_n]^\mathsf{T} \in\mathbb{R}^{n\times l}$ is the labeling confidence matrix, where $\mathbf{p}_i \in \mathbb{R}^l$ represents the labeling confidence vector of $\mathbf{x}_i$, and $p_{ij}$ denotes the probability of the $j$-th label being the ground-truth label of $\mathbf{x}_i$. For the base classifier utilizing the labeling confidence strategy, $\mathbf{P}$ is initialized according to the base classifier. Generally, it is initialized as follows:
\begin{equation}
p_{i j}=\left\{\begin{array}{ll}
\frac{1}{\sum_{j} y_{i j}} & \text { if } \quad y_{i j}=1 \\
0 & \text { otherwise }.
\end{array}\right.
\label{eq confidence initialize}
\end{equation}
The labeling confidence vector $\mathbf{p}_i$ typically satisfies the following constraints: $\sum_j p_{ij} = 1$, $0\leq p_{ij}\leq y_{ij}$ \cite{wang2021adaptivePLAGGD, 2018Leveraginglalo}. The first constrains $\mathbf{p}_i$ to be normalized, and the second indicates that only the confidence of a candidate label has a chance to be positive. It should be noted that the ideal state of $\mathbf{p}_i$ is one-hot.

Once the base classifier has been trained, its modeling output will be generated. Denote the modeling output matrix $\mathbf{M}=[\mathbf{m}_1,\mathbf{m}_2,...,\mathbf{m}_n]^\mathsf{T}\in\mathbb{R}^{n\times l}$, where $\mathbf{m}_{ij}$ represents the probability of the $j$-th label being $\mathbf{x}_i$'s ground-truth label as predicted by the base classifier. It should be noted that $\mathbf{m}_i$ may either be a real-valued or one-hot vector depending on the specific base classifier. For instance, for SURE \cite{feng2019partialsssfagdsg} $\mathbf{m}_i$ is real-valued, while for PL-KNN \cite{hullermeier2006learningPL-KNN} it is one-hot. Afterwards, $\mathbf{P}$ is updated to $\mathbf{P}_1$ through the following equation: 
\begin{equation}
    \mathbf{P}_1 =\mathcal{T}_0 \left(\mathcal{T}_{\mathbf{Y}}\left(\alpha  \mathbf{P} + (1-\alpha) \mathbf{M}\right)\right),
\end{equation}
where $\alpha$ is a hyper-parameter controlling the smoothness of the labeling confidence. $\mathcal{T}_0$, $\mathcal{T}_{{\mathbf{Y}}}$ are two thresholding operators in element-wise, i.e., $\mathcal{T}_0(a):=\max\{0,a\}$ with $a$ being a scalar and $\mathcal{T}_{{\mathbf{Y}}}(a):=\min\{{y}_{ij}, a\}$.

\subsection{Blurring Mechanism}

In the next step, a blurring mechanism is designed to further process the labeling confidence matrix $\mathbf{P}_1$ to $\mathbf{O}_1 \in \mathbb{R}^{n\times l}$ :
\begin{equation}
    \begin{aligned}
        \mathbf{Q}_1 =& \mathcal{E}( e^k\mathbf{P}_1) \odot \mathbf{Y},\\
        [\mathbf{O}_1]_i &= \frac{[\mathbf{Q}_1]_i}{\|[\mathbf{Q}_1]_i\|},
    \end{aligned}
\label{eq: Q blur}
\end{equation}
where $\mathcal{E}(\cdot)$ is an element-wise operator, for a matrix $\mathbf{A}=[a_{ij}]_{n\times l}\in\mathbb{R}^{n\times l}$, $\mathcal{E}(\mathbf{A}) = [{\rm exp}(a_{ij})]_{n\times l}$. $k$ is a temperature parameter that controls the extent of blurring of labeling confidence. $\odot$ represents the Hadamard product of two matrices. For matrices $\mathbf{A}$ and $\mathbf{B}$ with same size $n\times l$, $\mathbf{A} \odot \mathbf{B} = [a_{ij}b_{ij}]_{n\times l}$. The Hadamard product allows only the candidate label has a positive confidence. We then normalize each row of $\mathbf{Q}_1$ and output the result $\mathbf{O}_1$.

The labeling confidences are expected to be blurred at each stage of mutual supervision to prevent being overconfident in some false positive labels, hence we set $k<0$ and then normalize the results, which means two labeling confidences that differ significantly can also become close. For more details on why the blurring mechanism can blur the outputs, please refer to Sec. \ref{sec: blur ana}.

\subsection{Partner Classifier}
Since the partner classifier has significant impacts on the success of PLCP, designing an appropriate partner classifier is quite important. In order to better assist the base classifier, a classifier that specifies the labels that should not be assigned to a sample is instantiated as the partner classifier, since non-candidate label information is exactly accurate and opposite to the candidate label information, making it a valuable complement to the base classifier.

Denote the non-candidate label matrix $\hat{\mathbf{Y}} = [\hat{y}_{ij}]_{n\times l}$ where $\hat{y}_{ij} = 0$ (resp. $\hat{y}_{ij} = 1$) if the $j$-th label is (resp. is not) in the candidate label set of $\mathbf{x}_i$, and $\hat{\mathbf{P}}=[\hat{\mathbf{p}}_1,\hat{\mathbf{p}}_2,...,\hat{\mathbf{p}}_n]^\mathsf{T}\in \mathbb{R}^{n\times l}$ the non-candidate labeling confidence matrix, where $\hat{p}_{ij}$ represents the probability of the $j$-th label NOT being the ground-truth label of $\mathbf{x}_i$. Similar to the labeling confidence, $\hat{\mathbf{P}}$ is also constrained with two constraints: $\sum_j \hat{{p}}_{ij} = l-1, \hat{y}_{ij} \leq \hat{p}_{ij} \leq 1$. The first term constrains the sum of the probability of each label being invalid is strictly $l-1$, and the second indicates that only candidate label has a chance to update its non-candidate labeling confidence while the others keep 1 (invalid labels). Note that the ideal state of  $\hat{\mathbf{p}}_i$ is ``zero-hot'', i.e., only one element in $\hat{\mathbf{p}}_i$ is 0 with others 1. $\hat{\mathbf{P}}$ is initialized as $\hat{\mathbf{Y}}$. Suppose $\hat{\mathbf{W}}=[\hat{\mathbf{w}}_1, \hat{\mathbf{w}}_2,...,\hat{\mathbf{w}}_q]^\mathsf{T}\in \mathbb{R}^{q \times l}$ is the weight matrix, the partner classifier is formulated as follows:
\begin{equation}
\begin{split}
\min_{\hat{\mathbf{W}},\hat{\mathbf{b}}, \mathbf{C}}  & \quad\left\|\mathbf{X}\hat{\mathbf{W}}+\mathbf{1}_n\hat{\mathbf{b}}^{\mathsf{T}}-\mathbf{C}\right\|_F^2 + \lambda\left\|\hat{\mathbf{W}}\right\|_F^2\\
{\rm s.t.} & \quad \hat{\mathbf{Y}} \leq \mathbf{C} \leq \mathbf{1}_{n\times l}, \mathbf{C}\mathbf{1}_l = (l-1)\mathbf{1}_n,
\end{split}
\label{eq: partner classifier linear}
\end{equation}
where $\hat{\mathbf{b}} = [\hat{b}_1,\hat{b}_2,...,\hat{b}_l]^\mathsf{T}\in \mathbb{R}^l$ is the bias term, $\mathbf{1}_n \in \mathbb{R}^n$ is an all one vectors, $\mathbf{1}_{n\times l}$ is an all one matrix with size $n\times l$ and $\lambda$ is a hyper-parameters trading off these terms. $\|\hat{\mathbf{W}}\|_F$ is the Frobenius norm of the weight matrix. $\mathbf{C}\in \mathbb{R}^{n\times l}$ represents non-candidate labeling confidence, which is a temporary variable only used for optimization in the partner classifier. By solving this optimization problem, the partner classifier learns to specify which labels should not be assigned to each sample, improving the overall performance of the mutual supervision process.

\subsection{The Collaborative Term}
Since a label is either ground-truth or not and based on the constraints on $\mathbf{p}_i$ and $\hat{\mathbf{p}}_i$, the smallest value of $\mathbf{p}_i^{\mathsf{T}}\hat{\mathbf{p}}_i$ is obtained when $\hat{\mathbf{p}}_i$ is zero-hot ($\mathbf{p}_i$ is one-hot). Therefore, a collaborative relationship between the outputs of the partner classifier and the prior given by the base classifier can be formulated, with the partner classifier becoming
\begin{align}
\min_{\hat{\mathbf{W}},\hat{\mathbf{b}}, \mathbf{C}}  & \quad\left\|\mathbf{X}\hat{\mathbf{W}}+\mathbf{1}_n\hat{\mathbf{b}}^{\mathsf{T}}-\mathbf{C}\right\|_F^2 + \gamma {\rm tr}\left(\mathbf{O}_1\mathbf{C}^{\mathsf{T}}\right)+ \lambda\left\|\hat{\mathbf{W}}\right\|_F^2 \nonumber\\
{\rm s.t.} & \quad \hat{\mathbf{Y}} \leq \mathbf{C} \leq \mathbf{1}_{n\times l}, \mathbf{C}\mathbf{1}_l = (l-1)\mathbf{1}_n,
\label{eq: partner classifier}
\end{align}
where $\gamma$ is a hyper-parameter. Different from candidate labels, the non-candidate labels are directly accessible and accurate, therefore the information learnt by the partner classifier is precise, which can effectively complement the base classifier.

The problem in Eq. (\ref{eq: partner classifier}) can be solved via an alternative and iterative manner in Sec. \ref{sec numerical solution}, which results in an optimal weight matrix $\hat{\mathbf{W}}$ and bias term $\hat{\mathbf{b}}$. The modeling output $\hat{\mathbf{M}}$ for the training data is
\begin{equation}
    \hat{\mathbf{M}} = \mathbf{X}\hat{\mathbf{W}}+\mathbf{1}_n\hat{\mathbf{b}}^{\mathsf{T}}.
    \label{eq: partner modeling output}
\end{equation}
Afterwards, the non-candidate labeling confidence $\hat{\mathbf{P}}$ is updated to $\hat{\mathbf{P}}_1$ following
\begin{equation}
    \hat{\mathbf{P}}_1 =\mathcal{T}_1 \left(\mathcal{T}_{\mathbf{\hat{Y}}}\left(\alpha  \hat{\mathbf{P}} + (1-\alpha) \hat{\mathbf{M}}\right)\right),
\end{equation}
where $\mathcal{T}_1$, $\mathcal{T}_{\hat{\mathbf{Y}}}$ are two thresholding operators in element-wise, i.e., $\mathcal{T}_1(m):=\min\{1,m\}$ with $m$ being a scalar and $\mathcal{T}_{\hat{\mathbf{Y}}}(m):=\max\{\hat{{y}}_{ij}, m\}$. The non-candidate labeling confidence can be further processed as $\hat{\mathbf{O}}_1$ via the blurring mechanism:
\begin{equation}
\begin{aligned}
    \hat{\mathbf{Q}}_1 =& \mathcal{E}( e^k(1-\hat{\mathbf{P}}_1)) \odot \mathbf{Y},\\
    [\hat{\mathbf{O}}_1]_i &= \frac{[\hat{\mathbf{Q}}_1]_i}{\|[\hat{\mathbf{Q}}_1]_i\|}.
\end{aligned}
    \label{eq: hatQ blur}
\end{equation}
For convenience, we transform the non-candidate labeling confidence into labeling confidence in advance, and finally $\hat{\mathbf{Q}}_1$ is normalized as $\hat{\mathbf{O}}_1$, which is the supervision of the base classifier in the next iteration.

\begin{algorithm}[tb]
   \caption{The pseudo code of PLCP}
   \label{alg:PLCP}
\begin{algorithmic}[1]
   \STATE {\bfseries Input:} A base classifier $\mathcal{B}$ and the partner classifier $\mathcal{P}$, partial label data $\mathcal{D}$, hyper-parameter $\alpha$, $\gamma$ and $\lambda$, temperature parameter $k$, the maximum number of iteration $iter$ and unseen sample $\mathbf{x}^{\ast}$.
   \STATE {\bfseries Output:} The predicted label $y^{\ast}$
   \STATE Initialize $\mathbf{P}_0$, $\hat{\mathbf{P}}_0$ and $i=1$. 
   \WHILE{i$\leq$iter}
   \STATE Train the base classifier supervised by $\hat{\mathbf{O}}_{i-1}$ and obtain the modeling output $\mathbf{M}_i$.
   \STATE $ \mathbf{P}_i =\mathcal{T}_0 \left(\mathcal{T}_{\mathbf{Y}}\left(\alpha  \mathbf{P}_{i-1} + (1-\alpha) \mathbf{M}_{i}\right)\right)$.
   \STATE $ \mathbf{Q}_i = \mathcal{E}( e^k\mathbf{P}_i) \odot \mathbf{Y}$.
   \STATE Normalize $\mathbf{Q}_i$ and return $\mathbf{O}_i$. 
   \STATE Train the partner classifier supervised by $\mathbf{O}_i$ and obtain the modeling output $\hat{\mathbf{M}}_i$ according to Eq. (\ref{eq: partner modeling output}).
   \STATE $\hat{\mathbf{P}}_i =\mathcal{T}_1 \left(\mathcal{T}_{\mathbf{\hat{Y}}}\left(\alpha \hat{\mathbf{P}}_{i-1} + (1-\alpha) \hat{\mathbf{M}}_{i}\right)\right)$.
   \STATE $\hat{\mathbf{Q}}_i = \mathcal{E}(e^k(1-\hat{\mathbf{P}}_i)) \odot \mathbf{Y}$.
   \STATE Normalize $\hat{\mathbf{Q}}_i$ and return $\hat{\mathbf{O}}_i$. 
   \IF{the stopping criterion is meet}
   \STATE {\bfseries break;}
   \ENDIF
   \STATE $i=i+1$.
   \ENDWHILE
   \STATE {\bfseries Return: The predicted label $\mathbf{y}^\ast$} according to Eq. (\ref{eq:final pre}).
\end{algorithmic}
\end{algorithm}

\subsection{Stopping Criterion}
Through continuous mutual supervision, the base classifier could identify and rectify mislabeled samples. This iterative process of mutual supervision persists for multiple iterations. The iteration ceases when there is a consistent correction of over 5\% of mislabeled samples across two consecutive iterations.

For an unseen sample $\mathbf{x}^\ast$, suppose the non-candidate labeling confidence vector predicted by the partner classifier in the last iteration is $\hat{\mathbf{p}}^{pt}$. The prediction $y^\ast$ of PLCP is 
\begin{equation}
    y^\ast = {\rm argmax}_i \quad (1 - \hat{{p}}^{pt}_i),
\label{eq:final pre}
\end{equation}
Rewrite $\mathbf{P}_0 = \mathbf{P}$, $\hat{\mathbf{P}}_0 = \hat{\mathbf{P}}$, $\mathbf{H}_1 =\mathbf{H}$, $\hat{\mathbf{H}}_1 = \hat{\mathbf{H}}$ and $\hat{\mathbf{O}}_0 = \mathbf{P}$, the complete procedure of PLCP is shown in Algorithm \ref{alg:PLCP}.

\subsection{Numerical Solution of the Partner Classifier}\label{sec numerical solution}
Th optimization problem in Eq. (\ref{eq: partner classifier}) has three variables with different constraints, therefore we can adopt an alternative and iterative optimization to solve it. 

\subsubsection{Update $\hat{\mathbf{W}}$ and $\hat{\mathbf{b}}$}

With $\mathbf{C}$ fixed, the problem w.r.t. $\hat{\mathbf{W}}$ and $\hat{\mathbf{b}}$ can be written as

\begin{equation}
\min_{\hat{\mathbf{W}},\hat{\mathbf{b}}}\quad \left\|\mathbf{X}\hat{\mathbf{W}}+\mathbf{1}_n\hat{\mathbf{b}}^{\mathsf{T}}-\mathbf{C}\right\|_F^2 + \lambda\left\|\hat{\mathbf{W}}\right\|_F^2,
\label{eq: W subproblem}
\end{equation}
which is a least square problem with the closed-form solution as
\begin{equation}
\begin{split}
    \hat{\mathbf{W}} & = \left(\mathbf{X}^{\mathsf{T}}\mathbf{X} + \lambda\mathbf{I}_{n\times n}\right )\mathbf{X}^{\mathsf{T}}\mathbf{C}, \\ 
    \hat{\mathbf{b}} &= \frac{1}{n} \left(\mathbf{C}^{\mathsf{T}} \mathbf{1}_n - \hat{\mathbf{W}}^{\mathsf{T}}\mathbf{X}^{\mathsf{T}}\mathbf{1}_n\right),
\end{split} 
\label{eq w solution}
\end{equation}
where $\mathbf{I}_{n\times n}$ is the identity matrix with the size $n\times n$.

\subsubsection{Update $\mathbf{C}$}
With $\hat{\mathbf{W}}$ and $\hat{\mathbf{b}}$ fixed, the $\mathbf{C}$-subproblem can be formulated as 

\begin{equation}
\begin{split}
\min_{\mathbf{C}}&\quad \left\|\mathbf{X}\hat{\mathbf{W}}+\mathbf{1}_n\hat{\mathbf{b}}^{\mathsf{T}}-\mathbf{C}\right\|_F^2  + \gamma{\rm tr}\left(\mathbf{O}_1 \mathbf{C}^{\mathsf{T}}\right)\\
{\rm s.t.} & \quad \hat{\mathbf{Y}} \leq \mathbf{C} \leq \mathbf{1}_{n\times l}, \mathbf{C}\mathbf{1}_l = (l-1)\mathbf{1}_n.
\end{split}
\label{eq C subproblem}
\end{equation}
For simplicity, $\mathbf{O}_1$ is written as $\mathbf{O}$ and $\mathbf{J} = \mathbf{X}\hat{\mathbf{W}}+\mathbf{1}_n\hat{\mathbf{b}}^{\mathsf{T}}$. Notice that each row of $\mathbf{C}$ is independent to other rows, therefore the problem in Eq. (\ref{eq C subproblem}) can be solved row by row:

\begin{equation}
\begin{split}
\min_{\mathbf{C}_i}&\quad \mathbf{C}_i^{\mathsf{T}}\mathbf{C}_i + \left(\gamma\mathbf{O}_i-2\mathbf{J}_i \right)^{\mathsf{T}}\mathbf{C}_i\\
{\rm s.t.} & \quad \hat{\mathbf{Y}}_i \leq \mathbf{C}_i \leq \mathbf{1}_{n}, \mathbf{C}_i\mathbf{1}_l = l-1.
\end{split}
\label{eq C subproblem i}
\end{equation}
The problem in Eq. (\ref{eq C subproblem i}) is a standard Quadratic Programming (QP) problem, which can be solved by off-the-shelf QP tools.

\subsection{Extensions of PLCP}
We can also extend PLCP to a kernel version which extends the feature map to a higher dimensional space or a deep-learning based version which enables deep-learning based methods involved in PLCP. 

\subsubsection{Kernel Extension}

The linear classifier in Eq. (\ref{eq: partner classifier}) may lack some ability to tackle the complex relationships among the data. Therefore, we extend it to a kernel version. Denote $\phi(\cdot):\mathbb{R}^q\rightarrow \mathbb{R}^h$ the feature mapping that maps the feature space to some higher dimension space with $h$ dimensions, then we can rewrite Eq. (\ref{eq: W subproblem}) as

\begin{equation}
\begin{split}
    \min_{\hat{\mathbf{W}}, \hat{\mathbf{b}}}\quad& \|\mathbf{Z} \|_F^2 + \lambda \|\hat{\mathbf{W}}\|_F^2\\
    {\rm s.t.}\quad& \mathbf{Z} = \mathbf{\Phi}\hat{\mathbf{W}}+\mathbf{1}_n\hat{\mathbf{b}}^\mathsf{T} -\mathbf{C},
    \end{split}
    \label{eq kernel extension}
\end{equation}
where $\mathbf{\Phi}= [\phi(\mathbf{x}_1), \phi(\mathbf{x}_2),...,\phi(\mathbf{x}_n)]$. The Lagrangian function of this problem is

\begin{equation}
\begin{split}
    &\mathcal{L}(\hat{\mathbf{W}},\hat{\mathbf{b}},\mathbf{Z},\mathbf{A}) = \|\mathbf{Z} \|_F^2 + \lambda \|\hat{\mathbf{W}}\|_F^2 \\& - {\rm tr}(\mathbf{A}^{\mathsf{T}}(\mathbf{\Phi}\hat{\mathbf{W}}+\mathbf{1}_n\hat{\mathbf{b}}^\mathsf{T} - \mathbf{C}-\mathbf{Z})),
\end{split}
\end{equation}
where $\mathbf{A} = [a_{ij}]_{n\times l}\in \mathbb{R}^{n\times l}$ stores the Lagrange multipliers. Following the KKT conditions: 

\begin{equation}
    \begin{split}
        &\frac{\partial \mathcal{L}}{\partial \hat{\mathbf{W}}} = 2\lambda\hat{\mathbf{W}} -\mathbf{\Phi}^{\mathsf{T}}\mathbf{A}=0, \frac{\partial \mathcal{L}}{\partial \hat{\mathbf{b}}} = \mathbf{A}^{\mathsf{T}}\mathbf{1}_n=0, \\
       & \frac{\partial \mathcal{L}}{\partial \mathbf{Z}} = 2\mathbf{Z}+ \mathbf{A}=0,   \frac{\partial \mathcal{L}}{\partial \mathbf{A}} = \mathbf{\Phi}\hat{\mathbf{W}}+\mathbf{1}_n\hat{\mathbf{b}}^\mathsf{T} - \mathbf{C}-\mathbf{Z}=0,
       \label{eq: KKT}
    \end{split}
\end{equation}
we have the corresponding solution:
\begin{equation}
\begin{split}
    &\hat{\mathbf{W}} = \frac{\mathbf{\Phi}^{\mathsf{T}}\mathbf{A}}{2\lambda}, \mathbf{A} = \left(\frac{1}{2\lambda} \mathbf{K} + \frac{1}{2}\mathbf{I}_{n\times n}\right)^{-1}\left(\mathbf{C} - \mathbf{1}_n\hat{\mathbf{b}}^{\mathsf{T}}\right),\\
    &\mathbf{s} =\mathbf{1}_n^{\mathsf{T}}\left(\frac{1}{2\lambda} \mathbf{K} + \frac{1}{2}\mathbf{I}_{n\times n}\right)^{-1}, \hat{\mathbf{b}}^{\mathsf{T}} = \frac{\mathbf{s}\mathbf{C}}{\mathbf{s1}_n},
\end{split}
    \label{eq KKT solution}
\end{equation}
where $\mathbf{K}=\mathbf{\Phi}\mathbf{\Phi}^{\mathsf{T}}$ is the kernel matrix with its element $k_{ij} = \phi(\mathbf{x}_i)^\mathsf{T}\phi(\mathbf{x}_j) = \mathcal{K}(\mathbf{x}_i,\mathbf{x}_j)$ based on the kernel function $\mathcal{K}(\cdot,\cdot)$. For PLCP, Gaussian function $\mathcal{K}(\mathbf{x}_i,\mathbf{x}_j) = {\rm exp}(-\frac{\|\mathbf{x}_i - \mathbf{x}_j\|_2^2}{2\sigma^2}$) is adopted as the kernel function with $\sigma$ to the average distance of all pairs of training samples. The modeling output is denoted by 

\begin{equation}
\hat{\mathbf{M}} =  \frac{1}{2\lambda}\mathbf{KA} + \mathbf{1}_n\hat{\mathbf{b}}^{\mathsf{T}}.
\label{eq Kernel modeling output}
\end{equation}

\subsubsection{Deep Learning Extension}

PLCP can be also extended to a deep-learning version to further demonstrate its universality and effectiveness. The model of the majority of the deep-learning based methods usually contains a feature encoder followed by a prediction head, which predicts the labeling confidence of each sample. Denote a model in $\mathcal{B}$ with such architecture $g(\cdot)$, specifically, an additional model $\hat{g}(\cdot)$ with the same architecture as $g(\cdot)$ is introduced as the partner classifier, which predicts the non-candidate labeling confidence of each sample. Given an example $\mathbf{x}$ and its candidate label vector $\mathbf{y}$, the non-candidate loss is defined as
\begin{equation}
     \mathcal{L}_{com} =  - \sum_{i=1}^l (1-y_i) \log(\hat{g}_i(\mathbf{x})).
    \label{eq cross entropy }
\end{equation}
Afterwards, a collaborative loss is designed to link the two models:
\begin{equation}
\mathcal{L}_{col} = \sum_{i=1}^l v_i\hat{v}_i.
    \label{eq colla loss}
\end{equation}
Here, $\mathbf{v} = [v_1,v_2,..., v_l]$ and $\hat{\mathbf{v}} =[\hat{v}_1,\hat{v}_2,...,\hat{v}_l]$ are two blurred predictions of $g(\mathbf{x})$ and $\hat{g}(\mathbf{x})$ respectively, where
\begin{equation}
\begin{split}
    \mathbf{v} &= \frac{\mathcal{E}(e^k g(\mathbf{x})) \odot \mathbf{y}}{\|\mathcal{E}(e^k g(\mathbf{x})) \odot \mathbf{y}\|} \\
    \hat{\mathbf{v}} & = \mathbf{1}_l^{\mathsf{T}} - \frac{\mathcal{E}(e^k (1-\hat{g}(\mathbf{x}))) \odot \mathbf{y}}{\|\mathcal{E}(e^k (1-\hat{g}(\mathbf{x}))) \odot \mathbf{y}\|}.
\end{split}
    \label{eq blur loss}
\end{equation}
The overall loss function is:
\begin{equation}
    \mathcal{L} = \mathcal{L}_{ori} + \mathcal{L}_{com} + \mu \mathcal{L}_{col},
    \label{eq loss function}
\end{equation}
where $\mathcal{L}_{ori}$ is the original loss function of $\mathcal{B}$ and $\mu$ is a hyper-parameter.

\section{Experiments}\label{sec: exp}
\subsection{Compared Approaches}
To evaluate the effectiveness of PLCP, we couple it with several well-established partial label learning approaches. Suppose $\mathcal{B}$ represents any partial label learning classifier (i.e., the base classifier) and $\mathcal{B}$-PLCP is the $\mathcal{B}$ coupled with PLCP, the performances of $\mathcal{B}$ and $\mathcal{B}$-PLCP are compared to verify the effectiveness of PLCP. In this paper, $\mathcal{B}$ is instantiated by six stand-alone (non-deep) approaches, PL-CL\cite{jia2023complementary}, PL-AGGD \cite{wang2021adaptivePLAGGD}, SURE \cite{feng2019partialsssfagdsg}, LALO \cite{2018Leveraginglalo}, PL-SVM \cite{nguyen2008classificationplsvm} and PL-KNN \cite{hullermeier2006learningPL-KNN}, and two deep-learning based methods PICO \cite{wang2022pico} and PRODEN \cite{lv2020progressiveproden}. The hyper-parameters of $\mathcal{B}$ are all set according to the original papers.

\begin{table}[t]\scriptsize
\setlength{\tabcolsep}{2mm}
\renewcommand\arraystretch{1.0} 
    \caption{Characteristics of the real-world data sets.}

    \centering
    \begin{tabular}{c c c c c}
     \hline \hline
        Data set & \#examples & \#features & \#labels & average \#labels\\\hline
      
        FG-NET \cite{panis2016overviewFg-net} & 1002 & 262 & 78 & 7.48\\
         Lost \cite{cour2009learningLost} & 1122 & 108 & 16 & 2.23\\
        MSRCv2 \cite{liu2012conditional} & 1758 & 48 & 23 & 3.16 \\
        Mirflickr \cite{huiskes2008mirMirflickr} & 2780 & 1536 & 14 & 2.76 \\
        Soccer Player \cite{zeng2013learningSoccerplayer} & 17472 & 279 & 171 & 2.09\\
        Yahoo!News \cite{guillaumin2010multipleYahoonews} & 22991 & 163 & 219 & 1.91 \\
        \hline\hline
    \end{tabular}
    \label{tab:real-world data character}
\end{table}

\begin{table*}[t!]\scriptsize
\setlength{\tabcolsep}{1.5mm}
\renewcommand\arraystretch{1} 
    \centering
    \caption{Classification accuracy of each compared approach on the real-world data sets. For any compared approach $\mathcal{B}$, $\bullet$/$\circ$ indicates whether $\mathcal{B}$-PLCP is statistically superior/inferior to $\mathcal{B}$ according to pairwise $t$-test at significance level of 0.05.}
    \begin{tabular}{l l l l l l l l l}\hline\hline
         \multirow{2}{*}{Approaches} & \multicolumn{8}{c}{Data set} \\\cline{2-9} &\multicolumn{1}{c}{FG-NET} & \multicolumn{1}{c}{Lost} & \multicolumn{1}{c}{MSRCv2} & \multicolumn{1}{c}{Mirflickr} & \multicolumn{1}{c}{Soccer Player} &
         \multicolumn{1}{c}{Yahoo!News}&
         \multicolumn{1}{c}{FG-NET(MAE3)}&
         \multicolumn{1}{c}{FG-NET(MAE5)}\\\hline

        PL-CL & 0.072 $\pm$ 0.009 & 0.710 $\pm$ 0.022 & 0.469 $\pm$ 0.016 & 0.647 $\pm$ 0.012 & 0.534 $\pm$ 0.004 & 0.618 $\pm$ 0.003 & 0.433 $\pm$ 0.022 & 0.575 $\pm$ 0.015\\
        PL-CL-PLCP & 0.081 $\pm$ 0.006 $\bullet$ & 0.762 $\pm$ 0.017 $\bullet$ & 0.496 $\pm$ 0.010 $\bullet$ & 0.667 $\pm$ 0.010 $\bullet$ & 0.545 $\pm$ 0.001 $\bullet$ & 0.628 $\pm$ 0.002 $\bullet$ & 0.450 $\pm$ 0.013 $\bullet$ & 0.595 $\pm$ 0.006 $\bullet$ \\\hline
         
         PL-AGGD & 0.063 $\pm$ 0.010 & 0.690 $\pm$ 0.020 & 0.451 $\pm$ 0.023 & 0.610 $\pm$ 0.012  & 0.521 $\pm$ 0.004 & 0.605 $\pm$ 0.002 & 0.418 $\pm$ 0.020 & 0.562 $\pm$ 0.020 \\ 
        
         PL-AGGD-PLCP & 0.076 $\pm$ 0.009 $\bullet$ & 0.717 $\pm$ 0.020 $\bullet$ & 0.474 $\pm$ 0.015 $\bullet$ & 0.668 $\pm$ 0.014 $\bullet$ & 0.534 $\pm$ 0.003 $\bullet$ & 0.611 $\pm$ 0.002  $\bullet$ & 0.442 $\pm$ 0.018 $\bullet$ & 0.582 $\pm$ 0.011 $\bullet$ \\\hline
         
         SURE & 0.052 $\pm$ 0.007 & 0.709 $\pm$ 0.022  & 0.445 $\pm$ 0.022 & 0.630 $\pm$ 0.022 & 0.519 $\pm$ 0.004 & 0.598 $\pm$ 0.002 & 0.356 $\pm$ 0.020 & 0.494 $\pm$ 0.021  \\
        
         SURE-PLCP & 0.077 $\pm$ 0.009 $\bullet$ & 0.719 $\pm$ 0.021 $\bullet$ & 0.460 $\pm$ 0.018 $\bullet$ & 0.657 $\pm$ 0.020 $\bullet$ & 0.526 $\pm$ 0.003 $\bullet$ & 0.608 $\pm$ 0.002 $\bullet$ & 0.443 $\pm$ 0.015 $\bullet$ & 0.584 $\pm$ 0.012 $\bullet$ \\\hline
        
         LALO & 0.065 $\pm$ 0.010 & 0.682 $\pm$ 0.019 & 0.449 $\pm$ 0.016 & 0.629 $\pm$ 0.016 & 0.523 $\pm$ 0.003 & 0.601 $\pm$ 0.003 & 0.422 $\pm$ 0.023 & 0.566 $\pm$ 0.015 \\

         LALO-PLCP & 0.076 $\pm$ 0.010 $\bullet$ & 0.703 $\pm$ 0.017 $\bullet$ & 0.456 $\pm$ 0.018 $\bullet$ & 0.647 $\pm$ 0.018 $\bullet$ & 0.529 $\pm$ 0.004 $\bullet$ & 0.605 $\pm$ 0.002 $\bullet$ & 0.444 $\pm$ 0.015 $\bullet$ & 0.585 $\pm$ 0.015 $\bullet$ \\\hline

         PL-SVM & 0.043 $\pm$ 0.008 & 0.406 $\pm$ 0.033 & 0.389 $\pm$ 0.029 & 0.516 $\pm$ 0.022 & 0.412 $\pm$ 0.006 & 0.509 $\pm$ 0.006 & 0.314 $\pm$ 0.019 & 0.445 $\pm$ 0.016 \\
         
         PL-SVM-PLCP & 0.081 $\pm$ 0.010 $\bullet$ & 0.692 $\pm$ 0.024 $\bullet$ & 0.468 $\pm$ 0.021 $\bullet$ & 0.611 $\pm$ 0.018 $\bullet$ & 0.526 $\pm$ 0.006 $\bullet$ & 0.609 $\pm$ 0.002 $\bullet$ & 0.439 $\pm$ 0.017 $\bullet$ & 0.584 $\pm$ 0.015 $\bullet$ \\\hline
         
         PL-KNN & 0.036 $\pm$ 0.006 & 0.300 $\pm$ 0.018 & 0.393 $\pm$ 0.014 & 0.454 $\pm$ 0.016 & 0.492 $\pm$ 0.003 & 0.368 $\pm$ 0.004 & 0.288 $\pm$ 0.014 & 0.440 $\pm$ 0.017\\
         
         PL-KNN-PLCP & 0.077 $\pm$ 0.006 $\bullet$ & 0.668 $\pm$ 0.019 $\bullet$ & 0.469 $\pm$ 0.016 $\bullet$ & 0.608 $\pm$ 0.026 $\bullet$ & 0.523 $\pm$ 0.003 $\bullet$ & 0.597 $\pm$ 0.002 $\bullet$ & 0.434 $\pm$ 0.020 $\bullet$ & 0.578 $\pm$ 0.014 $\bullet$ \\\hline
         \multicolumn{9}{c}{Average Improvement Ratio:\quad PL-CL: 3.61\% \quad  PL-AGGD: 5.10 \% \quad SURE: 12.24 \%  \quad LALO: 4.01 \% \quad PL-SVM: 39.26 \% \quad PL-KNN: 53.98 \%}\\
         \hline\hline
    \end{tabular}        
    \label{tab:real world statistic}
\end{table*}
 
\begin{table*}[t!]\scriptsize
\setlength{\tabcolsep}{3.8mm}
\renewcommand\arraystretch{1} 
    \centering
    \caption{Classification accuracy of each compared approach on CIFAR-10 and CIFAR-100. For any compared approach $\mathcal{B}$, $\bullet$/$\circ$ indicates whether $\mathcal{B}$-PLCP is statistically superior/inferior to $\mathcal{B}$ according to pairwise $t$-test at significance level of 0.05.}
    \begin{tabular}{l l l l l l l }\hline\hline
         \multirow{2}{*}{Approaches} & \multicolumn{3}{c}{CIFAR-10} & \multicolumn{3}{c}{CIFAR-100}\\\cline{2-7} &\multicolumn{1}{c}{ $q=0.1$} & \multicolumn{1}{c}{$q=0.3$} & \multicolumn{1}{c}{$q=0.5$} & \multicolumn{1}{c}{$q=0.01$} & \multicolumn{1}{c}{$q=0.05$} &
         \multicolumn{1}{c}{$q=0.1$}\\\hline
        PICO & 94.39 $\pm$ 0.18 \% & 94.18 $\pm$ 0.12 \% & 93.58 $\pm$ 0.06 \% & 73.09 $\pm$ 0.34 \% & 72.74 $\pm$ 0.30 \% & 69.91 $\pm$ 0.24 \%\\
        PICO-PLCP & 94.80 $\pm$ 0.07 \% $\bullet$ & 94.53 $\pm$ 0.10 \% $\bullet$ & 93.67 $\pm$ 0.16 \% $\bullet$ & 73.90 $\pm$ 0.20 \% $\bullet$ & 73.51 $\pm$ 0.21 \% $\bullet$ & 70.00 $\pm$ 0.35 \%   \\ \rowcolor{gray!20}
        Fully Supervised  & \multicolumn{3}{c}{$\mathcal{B}$:\quad 94.91 $\pm$ 0.07 \% \quad $\mathcal{B}$-PLCP:\quad 95.02 $\pm$ 0.03 \%}  & \multicolumn{3}{c}{$\mathcal{B}$:\quad73.56 $\pm$ 0.10 \% \quad $\mathcal{B}$-PLCP: \quad 73.69 $\pm$ 0.14 \%} \\ \hline
        
        PRODEN & 89.12 $\pm$ 0.12 \% & 87.56 $\pm$ 0.15 \% & 84.92 $\pm$ 0.31 \% & 63.36 $\pm$ 0.33 \% & 60.88 $\pm$ 0.35 \% & 50.98 $\pm$ 0.74 \% \\
        PRODEN-PLCP & 89.63 $\pm$ 0.15 \% $\bullet$ & 88.19 $\pm$ 0.19 \% $\bullet$ & 85.31 $\pm$ 0.31 \% $\bullet$ & 64.20 $\pm$ 0.25 \% $\bullet$ & 61.78 $\pm$ 0.29 $\bullet$ & 50.76 $\pm$ 0.90 \% \\ \rowcolor{gray!20}
        Fully Supervised & \multicolumn{3}{c}{$\mathcal{B}$: \quad90.03 $\pm$ 0.13 \%\quad $\mathcal{B}$-PLCP:\quad 90.30 $\pm$ 0.08 \%}  & \multicolumn{3}{c}{$\mathcal{B}$:\quad65.03 $\pm$ 0.35 \%\quad $\mathcal{B}$-PLCP:\quad 65.52 $\pm$ 0.32 \%}\\
        
         \hline\hline
    \end{tabular}        
    \label{tab:deep-learning statistic}
\end{table*}

\begin{table*}[t!]\scriptsize
\setlength{\tabcolsep}{1.1mm}
\renewcommand\arraystretch{1} 
    \centering
     \caption{Ablation study of PLCP coupled with PL-AGGD on the real-world data sets. $\bullet$/$\circ$ indicates whether PL-AGGD-PLCP is statistically superior/inferior to its degenerated version according to pairwise $t$-test at significance level of 0.05.}
    \begin{tabular}{c c c l l l l l l l l}\hline\hline
         \multirow{2}{*}{Kernel} \setlength\tabcolsep{1pt} & \multirow{2}{*}{Partner} \setlength\tabcolsep{1pt}& \multirow{2}{*}{Blur} & \multicolumn{7}{c}{Data set} \\\cline{4-11} & & &\multicolumn{1}{c}{FG-NET} & \multicolumn{1}{c}{Lost} & \multicolumn{1}{c}{MSRCv2} & \multicolumn{1}{c}{Mirflickr} & \multicolumn{1}{c}{Soccer Player} &
         \multicolumn{1}{c}{Yahoo!News}&
         \multicolumn{1}{c}{FG-NET(MAE3)}&
         \multicolumn{1}{c}{FG-NET(MAE5)}\\\hline
         
        \multicolumn{3}{c}{PL-AGGD} & 0.063 $\pm$ 0.010 & 0.690 $\pm$ 0.020 & 0.451 $\pm$ 0.023 & 0.610 $\pm$ 0.012  & 0.521 $\pm$ 0.004 & 0.605 $\pm$ 0.002 & 0.418 $\pm$ 0.020 & 0.562 $\pm$ 0.020 \\\hline
        
        \scalebox{0.85}[1]{$\times$} & $P$ & \scalebox{0.85}[1]{$\times$} & 0.073 $\pm$ 0.011 $\bullet$ & 0.698 $\pm$ 0.023 $\bullet$ & 0.380 $\pm$ 0.013 $\bullet$ & 0.542 $\pm$ 0.013 $\bullet$ & 0.492 $\pm$ 0.003 $\bullet$ & 0.463 $\pm$ 0.002 $\bullet$ & 0.421 $\pm$ 0.020 $\bullet$ & 0.560 $\pm$ 0.016 $\bullet$ \\

        \checkmark & $P$ & \scalebox{0.85}[1]{$\times$} & 0.073 $\pm$ 0.006 $\bullet$ & 0.721 $\pm$ 0.024 $\circ$ & 0.471 $\pm$ 0.016 $\bullet$ & 0.664 $\pm$ 0.012 $\bullet$ & 0.521 $\pm$ 0.004 $\bullet$ & 0.608 $\pm$ 0.003 $\bullet$ & 0.422 $\pm$ 0.030 $\bullet$ & 0.566 $\pm$ 0.020 $\bullet$\\ 

        \checkmark & $O$ & \checkmark  & 0.071 $\pm$ 0.001 $\bullet$ & 0.721 $\pm$ 0.004 $\circ$ & 0.470 $\pm$ 0,020 $\bullet$ & 0.663 $\pm$ 0.011 $\bullet$ & 0.522 $\pm$ 0.003 $\bullet$ & 0.605 $\pm$ 0.002 $\bullet$ & 0.417 $\pm$ 0.022 $\bullet$ & 0.576 $\pm$ 0.014 $\bullet$ \\\hline

         \checkmark & $P$ & \checkmark & 0.076 $\pm$ 0.009 & 0.717 $\pm$ 0.020 & 0.474 $\pm$ 0.015 & 0.668 $\pm$ 0.014  & 0.534 $\pm$ 0.003 & 0.611 $\pm$ 0.002 & 0.442 $\pm$ 0.018 & 0.582 $\pm$ 0.011 \\ 
         
         \hline\hline
         
    \end{tabular}       
    \label{tab:ablation study}
\end{table*}

\begin{table*}[t!]\scriptsize
\setlength{\tabcolsep}{3.7mm}
\renewcommand\arraystretch{1} 
    \centering
     \caption{Ablation study of PLCP coupled with PICO on CIFAR-10 and CIFAR-100. $\bullet$/$\circ$ indicates whether PICO-PLCP is statistically superior/inferior to its degenerated version according to pairwise $t$-test at significance level of 0.05.}
    \begin{tabular}{c c l l l l l l }\hline
         \multirow{2}{*}{Partner} &\multirow{2}{*}{Blur} & \multicolumn{3}{c}{CIFAR-10} & \multicolumn{3}{c}{CIFAR-100}\\\cline{3-8} & & \multicolumn{1}{c}{ $q=0.1$} & \multicolumn{1}{c}{$q=0.3$} & \multicolumn{1}{c}{$q=0.5$} & \multicolumn{1}{c}{$q=0.01$} & \multicolumn{1}{c}{$q=0.05$} &
         \multicolumn{1}{c}{$q=0.1$}\\\hline
         \multicolumn{2}{c}{PICO} & 94.39 $\pm$ 0.18 \% & 94.18 $\pm$ 0.12 \% & 93.58 $\pm$ 0.06 \% & 73.09 $\pm$ 0.34 \% & 72.74 $\pm$ 0.30 \% & 69.91 $\pm$ 0.24 \% \\\hline
         
         $P$ & \scalebox{0.85}[1]{$\times$} & 94.60 $\pm$ 0.09 \% $\bullet$ & 94.20 $\pm$ 0.11 \% $\bullet$ & 93.63 $\pm$ 0.09 \% $\bullet$ & 73.71 $\pm$ 0.28 \% $\bullet$ & 73.17 $\pm$ 0.22 \% $\bullet$ & 69.91 $\pm$ 0.18 \% $\bullet$ \\

        $O$ & \checkmark & 94.44 $\pm$ 0.10 \% $\bullet$ & 94.37 $\pm$ 0.11 \% $\bullet$ & 93.59 $\pm$ 0.13 \% $\bullet$ & 73.43 $\pm$ 0.24 \% $\bullet$ & 72.98 $\pm$ 0.24 \% $\bullet$ & 69.77 $\pm$ 0.33 \% $\bullet$\\\hline
         
        \multicolumn{2}{c}{PICO-PLCP} & 94.80 $\pm$ 0.07 \%  & 94.53 $\pm$ 0.10 \% & 93.67 $\pm$ 0.16 \% & 73.90 $\pm$ 0.20 \% & 73.51 $\pm$ 0.21 \% & 70.00 $\pm$ 0.35 \%  \\ \hline\hline
    \end{tabular}      
    \label{tab:ablation study pico}
\end{table*}

\subsection{Comparison with Stand-alone Methods}
\subsubsection{Experimental Settings}
\label{sec:Experimental_Settings}
For PLCP, we set $\lambda = 0.05$, $\alpha =0.5$, $\gamma =2$ and $k=-1$, and the maximum iteration of mutual supervision is set to 5. For PL-CL, PL-AGGD, SURE, LALO and PL-SVM, the kernel function is Gaussian function, which is the same as we adopt. Ten runs of 50\%/50\% random train/test splits are performed, and the average accuracy with standard deviation is represented for all $\mathcal{B}$ and $\mathcal{B}$-PLCP. For PL-SVM and PL-KNN which do not adopt labeling confidence strategy, $\hat{\mathbf{O}}$ is further processed as ${\mathcal{G}}(\hat{\mathbf{O}}-\mathbf{P})$ where ${\mathcal{G}}$ is an element-wise operator and $\mathcal{G}(x) = 1$ if $x\geq0$ otherwise 0 with $x$ being a scalar. 

We conduct experiments on six real-world partial label data sets collected from several domains and tasks, including FG-NET \cite{panis2016overviewFg-net} for facial age estimation, Lost \cite{cour2009learningLost}, Soccer Player \cite{zeng2013learningSoccerplayer} and Yahoo!News \cite{guillaumin2010multipleYahoonews} for automatic face naming, MSRCv2 \cite{liu2012conditional} for object classification and Mirflickr \cite{huiskes2008mirMirflickr} for web image classification. 
%For facial age estimation task, the ages annotated by crowd-sourced labelers are considered as each human face's candidate labels.
%For automatic face naming task, each face scratched from a video or an image is presented as a sample while the names extracted from the corresponding titles or captions are its candidate labels.
For object classification task, image segmentations are taken as instances with objects appearing in the same image as candidate labels. 
For web image classification task, each example consists of images and a set of candidate tags extracted from the web page. The details of the data sets are summarized in Table \ref{tab:real-world data character}.

\subsubsection{Performance on Real-World Data Set}
It is noteworthy that the number of average label of FG-NET is quite large, which could cause quite low classification accuracy for all approaches. The common strategy is to evaluate the mean absolute error (MAE) between the predicted age and the ground-truth one. Specifically, we add another two sets of comparisons on FG-NET w.r.t. MAE3/MAE5, which means that test samples can be considered to be correctly classified if the difference between the predicted age and true age is no more than 3/5 years. Table \ref{tab:real world statistic} summaries the classification accuracy with standard deviation of each approach on real-world data sets, where we can observe that

\begin{itemize}
    \item  $\mathcal{B}$-PLCP significantly outperforms the base classifier $\mathcal{B}$ in all cases according to the pairwise $t$-test with a significance level of 0.05, which validates the effectiveness of PLCP.
    \item State-of-the-art (SOTA) approaches, such as PL-CL, PL-AGGD, LALO and SURE, can also be significantly improved by PLCP on all data sets. For instance, on FG-NET the performance of SURE can be improved by \textbf{45\%}, and on FG-NET(MAE3) the performance of LALO can be improved by \textbf{5\%}. Additionally, on Lost PL-AGGD can be improved by \textbf{4\%} and PL-CL can be improved by \textbf{5.5\%}. The partner classifier's non-candidate label information effectively and significantly aids disambiguation, leading to outstanding performance of the PLCP framework.
    
    \item The performances of PL-SVM and PL-KNN are improved significantly and impressively when coupled with PLCP. For example, on FG-NET the performance of PL-KNN-PLCP is \textbf{more than two times} better than that of PL-KNN, and PL-SVM-PLCP's performance is \textbf{more than 1.5 times} better than that of PL-SVM on Lost. Although PL-SVM and PL-KNN are inferior to the SOTA methods, \textbf{they can achieve almost the same performance as SOTA's when coupled with PLCP}.

    \end{itemize}

\subsection{Comparison with Deep-learning Based Methods}
\subsubsection{Experimental Settings}

In PLCP, we set $k=-1$ and $\mu=0.5$. All the hyper-parameters and experimental setups of $\mathcal{B}$ are set based on their original papers.

We conduct experiments on two benchmarks CIFAR-10 and CIFAR-100 \cite{krizhevsky2009learning}, and following the settings in \cite{wang2022pico, lv2020progressiveproden}, we generate false positive labels by flipping negative labels $\hat{y} \neq y$ for each sample with a probability $q = P(\hat{y}\in \mathcal{Y}|\hat{y} \neq y)$, where $y$ is the ground-truth label. In other words, the probability of a false positive is uniform across all $l-1$ negative labels. We combine the flipped ones with the ground-truth label to create the set of candidate labels. Specifically, $q$ is set to 0.1, 0.3 and 0.5 for CIFAR-10 and 0.01, 0.05 and 0.1 for CIFAR-100. Five independent runs are performed with the the average accuracy and standard deviation recorded (with different seeds).

\subsubsection{Performance on CIFAR-10 and CIFAR-100}

Table \ref{tab:deep-learning statistic} presents the classification accuracy with standard deviation of each approach on CIFAR-10 and CIFAR-100, where we can find that 

\begin{itemize}
    \item $\mathcal{B}$-PLCP consistently outperforms the base classifier $\mathcal{B}$ in 83.3\% cases, which confirms that PLCP effectively improves the performance of deep-learning based models. Notably, $\mathcal{B}$-PLCP is never significantly outperformed by any $\mathcal{B}$.
    
    \item Although the performance improvement of $\mathcal{B}$-PLCP on different settings appears to be limited, it is important to note that its performance is very close to that of fully supervised $\mathcal{B}$. In some cases, it even surpasses the performance of fully supervised one.
    
\end{itemize}

\subsection{Ablation Study}
\subsubsection{Effectiveness of the Non-candidate Label Information}
In order to validate the effectiveness of the non-candidate label information, we instantiate the partner classifier with the form as 
\begin{equation}
\begin{split}
\min_{{\mathbf{W}},{\mathbf{b}}, \mathbf{L}}  & \quad\left\|\mathbf{X}{\mathbf{W}}+\mathbf{1}_n{\mathbf{b}}^{\mathsf{T}}-\mathbf{L}\right\|_F^2 + \lambda\left\|\hat{\mathbf{W}}\right\|_F^2\\
& \qquad \qquad + \gamma {\rm tr}\left(\mathbf{O}_1(\mathbf{1}_{n\times l} - \mathbf{L})^{\mathsf{T}}\right) \\
{\rm s.t.} & \quad \mathbf{0}_{n\times l} \leq \mathbf{L} \leq \mathbf{Y}, \mathbf{L}\mathbf{1}_l = \mathbf{1}_n,
\end{split}
\end{equation}
where $\mathbf{W}$ and $\mathbf{b}$ are two model parameters and $\mathbf{L}\in\mathbb{R}^{n\times l}$ is the labeling confidence matrix. Denote this classifier $O$ and the original partner classifier $P$, the classification accuracy of PL-AGGD mutual-supervised with $O$ on each real-world data set is shown in Table \ref{tab:ablation study}, and that of PICO on CIFAR data set is shown in Table \ref{tab:ablation study pico}. It is obvious that the performances of PL-AGGD and PICO coupled with $O$ are significantly inferior to that with $P$ in 87.5\% and all cases respectively, which validates the usefulness of the non-candidate label information. The accurate non-candidate label information provided by the partner classifier effectively complements the base classifier, leading to better performance.

\begin{figure*}[ht!]
  \centering
  
  \subfloat[\label{fig:sensitive-alpha}][Sensitivity w.r.t. $\lambda$.]{ \includegraphics[scale=0.28]{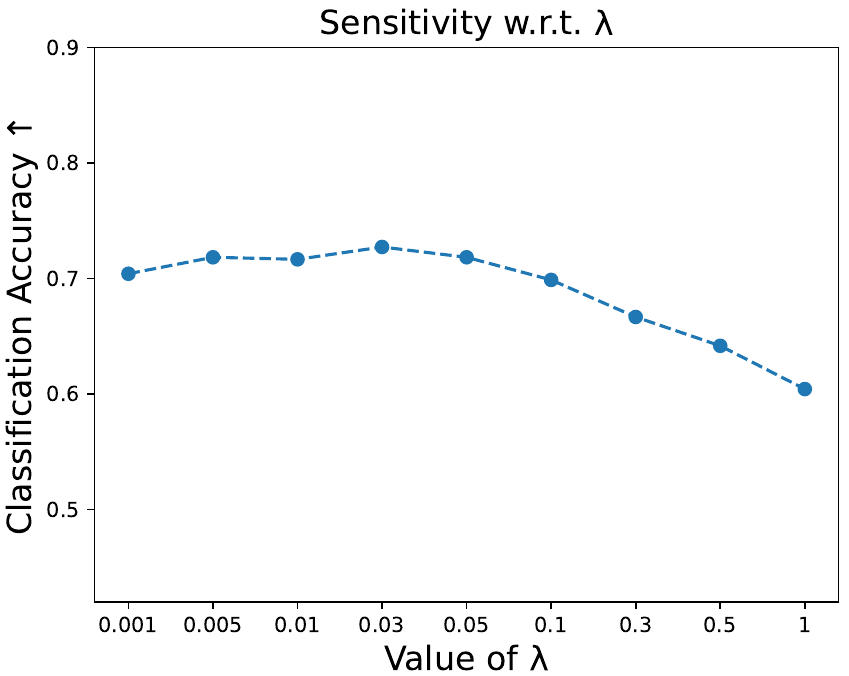}}
  \subfloat[\label{fig:sensitive-lambda}][Sensitivity w.r.t. $\alpha$.]{ \includegraphics[scale=0.28]{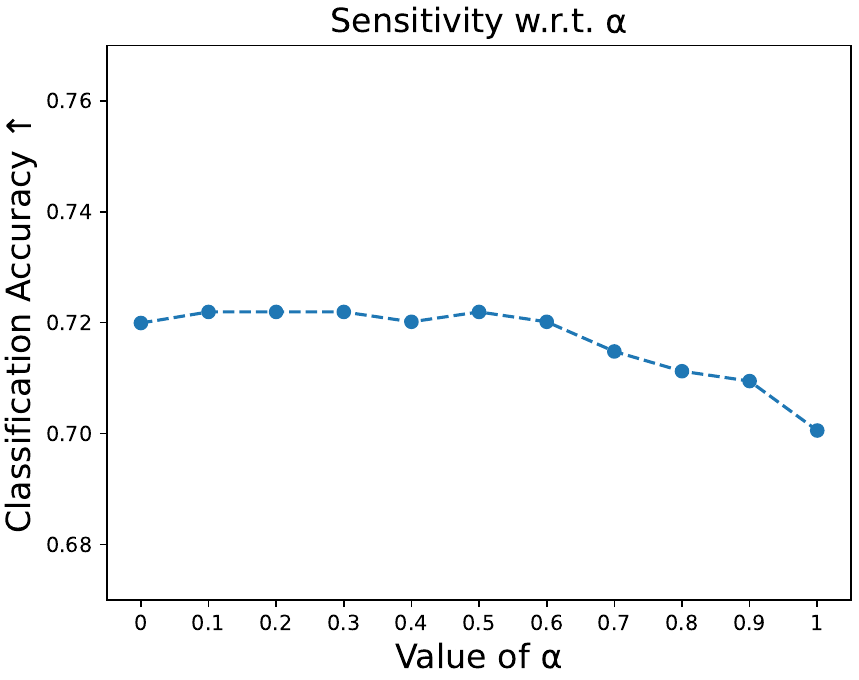}}
 \subfloat[\label{fig:sensitive-gamma}][Sensitivity w.r.t. $\gamma$.]{
 \includegraphics[scale=0.28]{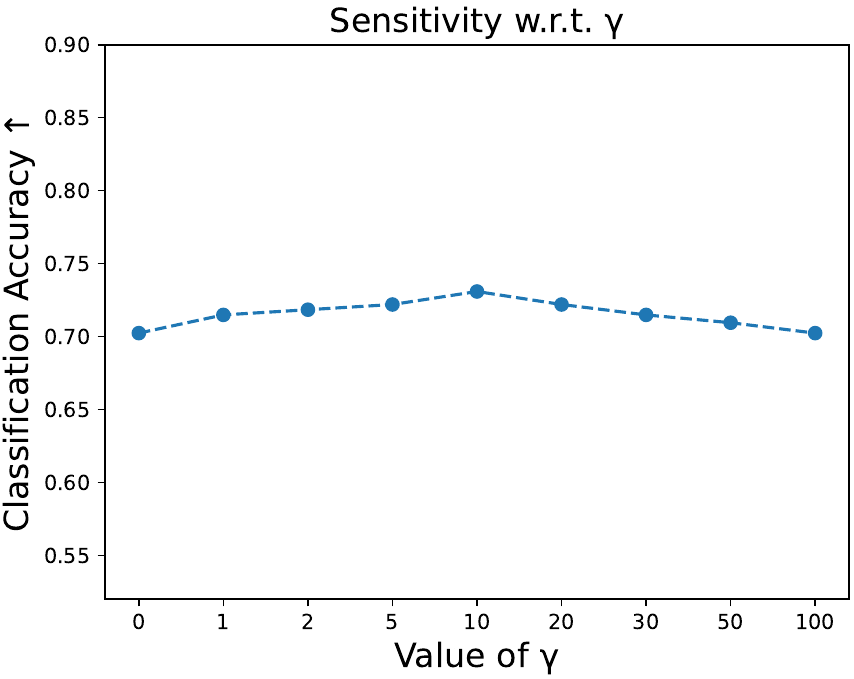}}
 \subfloat[\label{fig:sensitive-k}][Sensitivity w.r.t. $k$.]{ \includegraphics[scale=0.28]{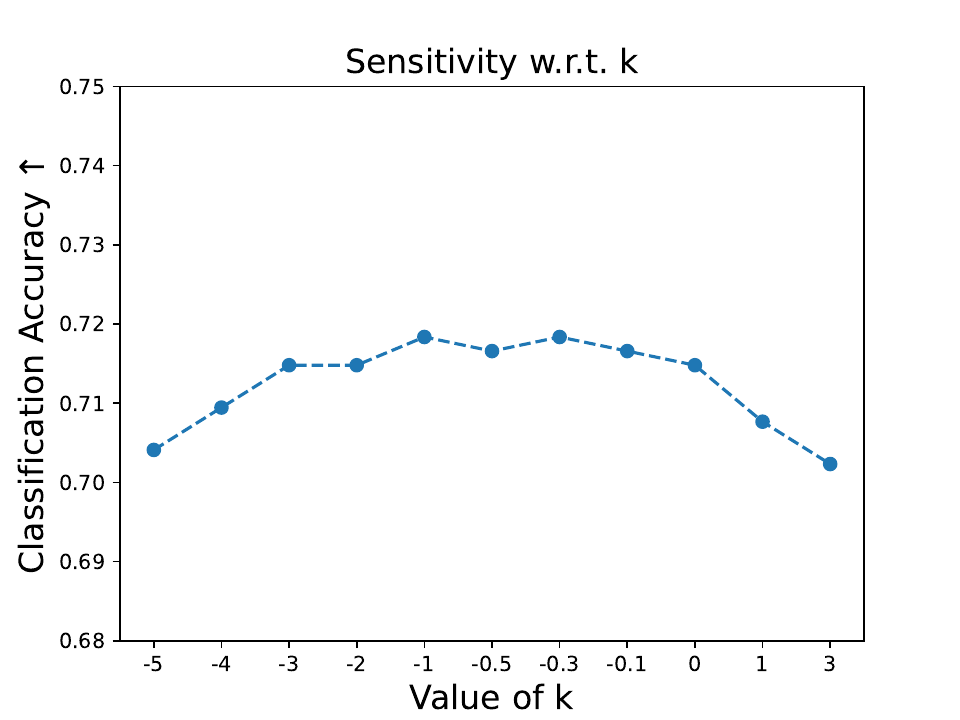}}
 
  \caption{Sensitivity of PLCP.}
  \label{fig:sensitive}
\end{figure*}

\begin{table*}[t!]\footnotesize
\setlength{\tabcolsep}{0.9mm}
\renewcommand\arraystretch{1} 
    \centering
    \caption{Correction ratio of each method on real-world data sets.}
    \begin{tabular}{l r r r r r r r r}\hline\hline
         \multirow{2}{*}{Approaches} & \multicolumn{8}{c}{Data set} \\\cline{2-9} &\multicolumn{1}{c}{FG-NET} & \multicolumn{1}{c}{Lost} & \multicolumn{1}{c}{MSRCv2}  & \multicolumn{1}{c}{Mirflickr}  & \multicolumn{1}{c}{Soccer Player} &
         \multicolumn{1}{c}{Yahoo!News}&
         \multicolumn{1}{c}{FG-NET(MAE3)}&
         \multicolumn{1}{c}{FG-NET(MAE5)}\\\hline
         PL-CL & 3.65 $\pm$ 0.02 \% & 21.29 $\pm$ 1.04 \% & 6.24 $\pm$ 0.48 \% & 7.96 $\pm$ 1.19 \% & 0.77 $\pm$ 0.08 \% & 4.19 $\pm$ 0.42 \% & 1.82 $\pm$ 0.19 \% & 2.46 $\pm$ 0.28 \% \\
         
         PL-AGGD & 3.64 $\pm$ 0.04 \% & 20.00 $\pm$ 1.55 \% & 5.17 $\pm$ 0.49 \% & 9.31 $\pm$ 1.35 \% & 1.39 $\pm$ 0.21 \% & 1.99 $\pm$ 0.13 \% & 3.92 $\pm$ 0.84 \% & 4.81 $\pm$ 1.02 \%\\
         
         SURE & 6.11 $\pm$ 0.02 \% & 6.67 $\pm$ 0.44 \% & 3.93 $\pm$ 0.73 \% & 11.39 $\pm$ 2.11 \% & 1.83 $\pm$ 0.26 \% & 7.45 $\pm$ 0.39 \% & 26.30 $\pm$ 2.48 \%& 34.90 $\pm$ 1.26 \%\\
         
         LALO & 3.46 $\pm$ 0.01 \% & 13.77 $\pm$ 1.72 \% & 1.94 $\pm$ 0.35 \% & 7.43 $\pm$ 0.91 \% & 1.41 $\pm$ 0.22 \% & 3.48 $\pm$ 0.19 \% & 4.80 $\pm$ 1.26 \% & 5.85 $\pm$ 0.23 \%\\
         
         PL-SVM & 7.49 $\pm$ 0.21 \% & 64.89 $\pm$ 2.51 \% & 31.28 $\pm$ 2.19 \% & 38.86 $\pm$ 2.47 \% & 28.10 $\pm$ 0.73 \% & 34.80 $\pm$ 0.88 \% & 38.98 $\pm$ 1.92 \% & 47.65 $\pm$ 2.59 \%\\
         
         PL-KNN & 8.42 $\pm$ 0.17 \% & 56.93 $\pm$ 1.93 \% & 34.46 $\pm$ 1.83 \% & 49.81 $\pm$ 2.19 \% & 6.81 $\pm$ 0.52 \% & 43.47 $\pm$ 0.81 \% & 41.74 $\pm$ 2.01 \% & 39.22 $\pm$ 1.82 \% \\
         \hline\hline
    \end{tabular}        
    \label{tab:appeal correction statistic}
\end{table*}

\begin{table*}[t!]\scriptsize
\setlength{\tabcolsep}{1.5mm}
\renewcommand\arraystretch{1} 
    \centering
    \caption{Transductive accuracy of each compared approach on the real-world data sets. For any compared approach $\mathcal{B}$, $\bullet$/$\circ$ indicates whether $\mathcal{B}$-PLCP is statistically superior/inferior to $\mathcal{B}$ according to pairwise $t$-test at significance level of 0.05.}
    \begin{tabular}{l l l l l l l l l }\hline\hline
         \multirow{2}{*}{Approaches} & \multicolumn{8}{c}{Data set} \\\cline{2-9} &\multicolumn{1}{c}{FG-NET} & \multicolumn{1}{c}{Lost} & \multicolumn{1}{c}{MSRCv2}  & \multicolumn{1}{c}{Mirflickr}  & \multicolumn{1}{c}{Soccer Player} &
         \multicolumn{1}{c}{Yahoo!News}&
         \multicolumn{1}{c}{FG-NET(MAE3)}&
         \multicolumn{1}{c}{FG-NET(MAE5)}\\\hline

         PL-CL & 0.159 $\pm$ 0.016 & 0.832 $\pm$ 0.019 & 0.585 $\pm$ 0.012 & 0.697 $\pm$ 0.019 & 0.715 $\pm$ 0.001 & 0.827 $\pm$ 0.003 & 0.600 $\pm$ 0.029 & 0.737 $\pm$ 0.018\\
         
         PL-CL-PLCP & 0.181 $\pm$ 0.012 $\bullet$ & 0.854 $\pm$ 0.010 $\bullet$ & 0.638 $\pm$ 0.008 $\bullet$ & 0.704 $\pm$ 0.021 $\bullet$ & 0.722 $\pm$ 0.004 $\bullet$ & 0.831 $\pm$ 0.003 $\bullet$ & 0.618 $\pm$ 0.023 $\bullet$ & 0.747 $\pm$ 0.018 $\bullet$ \\\hline

         PL-AGGD & 0.141 $\pm$ 0.012 & 0.793 $\pm$ 0.020 & 0.557 $\pm$ 0.015 & 0.695 $\pm$ 0.015 & 0.669 $\pm$ 0.003 & 0.808 $\pm$ 0.005 & 0.571 $\pm$ 0.027 & 0.709 $\pm$ 0.025 \\ 
        
         PL-AGGD-PLCP & 0.167 $\pm$ 0.015 $\bullet$ & 0.829 $\pm$ 0.016 $\bullet$ & 0.638 $\pm$ 0.011 $\bullet$ & 0.715 $\pm$ 0.015 $\bullet$ & 0.714 $\pm$ 0.004 $\bullet$ & 0.831 $\pm$ 0.004 $\bullet$ & 0.600 $\pm$ 0.022 $\bullet$ & 0.738 $\pm$ 0.016 $\bullet$ \\\hline
         
         SURE & 0.158 $\pm$ 0.012 & 0.796 $\pm$ 0.026 & 0.603 $\pm$ 0.016 & 0.650 $\pm$ 0.024 & 0.700 $\pm$ 0.003 & 0.798 $\pm$ 0.005 & 0.590 $\pm$ 0.019 & 0.727 $\pm$ 0.020\\
        
         SURE-PLCP & 0.170 $\pm$ 0.011 $\bullet$ & 0.836 $\pm$ 0.021 $\bullet$ & 0.622 $\pm$ 0.015 $\bullet$ & 0.697 $\pm$ 0.017 $\bullet$ & 0.706 $\pm$ 0.005 $\bullet$ & 0.827 $\pm$ 0.006 $\bullet$ & 0.605 $\pm$ 0.026 $\bullet$ & 0.739 $\pm$ 0.018 $\bullet$\\\hline
        
         LALO & 0.153 $\pm$ 0.017 & 0.818 $\pm$ 0.019 & 0.548 $\pm$ 0.009 & 0.681 $\pm$ 0.013 & 0.688 $\pm$ 0.004 & 0.822 $\pm$ 0.004 & 0.593 $\pm$ 0.025 & 0.730 $\pm$ 0.015\\

         LALO-PLCP & 0.168 $\pm$ 0.018 $\bullet$ & 0.831 $\pm$ 0.019 $\bullet$ & 0.620 $\pm$\ 0.009 $\bullet$ & 0.694 $\pm$ 0.019 $\bullet$ & 0.706 $\pm$ 0.004 $\bullet$ & 0.827 $\pm$ 0.004 $\bullet$ &  0.604 $\pm$ 0.025 $\bullet$ & 0.741 $\pm$ 0.019 $\bullet$ \\\hline

         PL-SVM & 0.176 $\pm$ 0.015 & 0.609 $\pm$ 0.055 & 0.570 $\pm$ 0.040 & 0.581 $\pm$ 0.022 & 0.660 $\pm$ 0.008 & 0.691 $\pm$ 0.005 & 0.566 $\pm$ 0.025 & 0.706 $\pm$ 0.024 \\
         
         PL-SVM-PLCP & 0.188 $\pm$ 0.008 $\bullet$ & 0.792 $\pm$ 0.036 $\bullet$ & 0.639 $\pm$ 0.027 $\bullet$ & 0.630 $\pm$ 0.028 $\bullet$ & 0.705 $\pm$ 0.003 $\bullet$ & 0.823 $\pm$ 0.004 $\bullet$ & 0.580 $\pm$ 0.027 $\bullet$ & 0.721 $\pm$ 0.019 $\bullet$ \\\hline
         
         PL-KNN & 0.041 $\pm$ 0.007 & 0.337 $\pm$ 0.030 & 0.415 $\pm$ 0.014 & 0.466 $\pm$ 0.013 & 0.493 $\pm$ 0.004 & 0.403 $\pm$ 0.010 & 0.285 $\pm$ 0.017 & 0.438 $\pm$ 0.015 \\
         
         PL-KNN-PLCP & 0.168 $\pm$ 0.012 $\bullet$ & 0.781 $\pm$ 0.033 $\bullet$ & 0.639 $\pm$ 0.021 $\bullet$ & 0.628 $\pm$ 0.015 $\bullet$ & 0.699 $\pm$ 0.006 $\bullet$ & 0.793 $\pm$ 0.004 $\bullet$ & 0.579 $\pm$ 0.025 $\bullet$ & 0.722 $\pm$ 0.018 $\bullet$ \\\hline\hline
    \end{tabular}        
    \label{tab:transductive statistic}
\end{table*}

\subsubsection{Usefulness of the Blurring Mechanism}
It is also interesting to investigate the usefulness of the blurring mechanism in PLCP. By simply normalizing $\mathbf{P}_1$ to $\mathbf{O}_1$ and $\hat{\mathbf{P}}_1$ to $\hat{\mathbf{O}}_1$ (i.e., skipping Eq. (\ref{eq: Q blur}) and Eq. (\ref{eq: hatQ blur})), the classification accuracy of PLCP without this mechanism is recorded in Table \ref{tab:ablation study}-\ref{tab:ablation study pico}, where we can find that the performance of PL-AGGD-PLCP and PICO-PLCP w/o Blur is significantly inferior to those with Blur in 87.5\% and all cases respectively. The blurring mechanism effectively tackles the overconfidence issue in PLL, leading to outstanding performance. 

Furthermore, we examine the impact of various values of $k$ on the performance of PLCP, as depicted in Fig. \ref{fig:sensitive}(d). It can be clearly observed that when $k$ is too small, the performance of PLCP deteriorates significantly. Additionally, the performance will be also exacerbated when $k\geq0$ (strictly $k>\ln{2}$), for in this case the predictions of the classifiers may be enhanced rather than being blurred, amplifying small differences between two values. As discussed in Sec. \ref{sec: colla ana}, PLCP contributes little to appeal and disambiguation in this situation, resulting in inferior performance.

\subsubsection{Influence of Kernel Extension}
We also conduct experiments to show the improvement of kernel extension used in partner classifier. Comparing the results in the first and second rows in Table \ref{tab:ablation study}, we can observe that the performance of PL-AGGD coupled with the classifier using kernel extension is superior to that without kernel extension on all the data sets, which validates the effectiveness of the kernel.

\subsubsection{Sensitive Analysis \label{sensitive section}}
Figs. \ref{fig:sensitive}(a)-(d) illustrate the sensitivity of PL-AGGD coupled with PLCP w.r.t $\lambda$, $\alpha$, $\gamma$ and $k$ on Lost, where we can find that PL-AGGD-PLCP performs quite stably with the hyper-parameters changing within a reasonably wide range, which is also a desirable property in practice.

\section{Further Analysis}\label{sec:fur}

\subsection{The Appeal Ability}
In order to evaluate the appeal ability of PLCP, we propose a metric called ``correction ratio''. Specifically, we record the indices of mislabeled samples on each data set, and ascertain how many of them can be successfully corrected by PLCP. The results, shown in Table \ref{tab:appeal correction statistic}, demonstrate that PLCP adeptly addresses mislabeled samples, successfully identifying and correcting them with a high degree of probability. While it is conceivable that PLCP could result in mislabeling of originally correct samples, we have observed that such phenomenon occurs with a relatively low probability of 2.56\%.

\subsection{Improvement of the Disambiguation Ability}
Transductive accuracy (i.e., classification accuracy on training samples) reflects the disambiguation ability of a PLL approach \cite{wang2021adaptivePLAGGD, cour2011learning, zhang2016partial23232}. In order to validate whether PLCP can correct some mislabeled samples and truly improve the disambiguation ability of $\mathcal{B}$, we summarize the transductive accuracy of $\mathcal{B}$ and $\mathcal{B}$-PLCP on real-world data sets and the results are shown in Table \ref{tab:transductive statistic}. It is obvious that $\mathcal{B}$-PLCP outperforms $\mathcal{B}$ in all cases according to the pairwise $t$-test with a significance level of 0.05, which validates that the disambiguation ability of $\mathcal{B}$ can be truly improved by PLCP. In other words, by integrating PLCP, classifiers can better identify and correct mislabeled samples, leading to outstanding performance.

\begin{table*}[t!]\scriptsize
\setlength{\tabcolsep}{1.3mm}
\renewcommand\arraystretch{1} 
    \centering
    \caption{Comparison of aggressive term and collaborative term of PLCP coupled with PL-AGGD on real-world data sets. PL-AGGD-PLCP(A) represents the one using the aggressive term. $\bullet$/$\circ$ indicates whether PL-AGGD-PLCP is statistically superior/inferior to PL-AGGD-PLCP(A) according to pairwise $t$-test at significance level of 0.05.}
    \begin{tabular}{l l l l l l l l l }\hline\hline
         \multirow{2}{*}{Approaches} & \multicolumn{8}{c}{Data set} \\\cline{2-9} &\multicolumn{1}{c}{FG-NET} & \multicolumn{1}{c}{Lost} & \multicolumn{1}{c}{MSRCv2}  & \multicolumn{1}{c}{Mirflickr}  & \multicolumn{1}{c}{Soccer Player} &
         \multicolumn{1}{c}{Yahoo!News}&
         \multicolumn{1}{c}{FG-NET(MAE3)}&
         \multicolumn{1}{c}{FG-NET(MAE5)}\\\hline
       PL-AGGD & 0.063 $\pm$ 0.010 & 0.690 $\pm$ 0.020 & 0.451 $\pm$ 0.023 & 0.610 $\pm$ 0.012  & 0.521 $\pm$ 0.004 & 0.605 $\pm$ 0.002 & 0.418 $\pm$ 0.020 & 0.562 $\pm$ 0.020 \\ 
       
        PL-AGGD-PLCP(A) & 0.070 $\pm$ 0.010 & 0.697 $\pm$ 0.019 & 0.468 $\pm$ 0.022 & 0.657 $\pm$ 0.009 & 0.522 $\pm$ 0.004 & 0.604 $\pm$ 0.002 & 0.426 $\pm$ 0.019 & 0.564 $\pm$ 0.016\\
        
         PL-AGGD-PLCP & 0.076 $\pm$ 0.009 $\bullet$ & 0.717 $\pm$ 0.020 $\bullet$ & 0.474 $\pm$ 0.015 $\bullet$ & 0.668 $\pm$ 0.014 $\bullet$ & 0.534 $\pm$ 0.003 $\bullet$ & 0.611 $\pm$ 0.002  $\bullet$ & 0.442 $\pm$ 0.018 $\bullet$ & 0.582 $\pm$ 0.011 $\bullet$ \\ \hline\hline
    \end{tabular}        
    \label{tab:aggressive}
\end{table*}

\subsection{Analysis of the Collaborative Term \label{sec: colla ana}}
\subsubsection{How dose the collaborative term work?}
We focus our attention on the following convex optimization problem:
\begin{equation}
\begin{split}
\min_{\mathbf{C}} & \quad{\rm tr}(\mathbf{O}\mathbf{C}^{\mathsf{T}}) \\
{\rm s.t.} & \quad \hat{\mathbf{Y}} \leq \mathbf{C} \leq \mathbf{1}_{n\times l}, \mathbf{C}\mathbf{1}_l = (l-1)\mathbf{1}_n,
\end{split}
    \label{eq therotical tr question}
\end{equation}
where $\mathbf{O}$ is the blurred labeling confidence matrix. Since each row of $\mathbf{C}$ and $\mathbf{O}$ is independent to each other, the problem in Eq. (\ref{eq therotical tr question}) can be further transformed into a series of subproblems, which can be rewritten as
\begin{equation}
\begin{split}
\min_{\mathbf{C}_i} & \quad L =\sum_{j} O_{ij}C_{ij} \\
{\rm s.t.} & \quad \hat{y}_{ij} \leq {C}_{ij} \leq 1, \sum_{j} C_{ij}=l-1, \\ & \quad j=1,2,...,l.
\end{split}
    \label{eq therotical tr question ij}
\end{equation}
For simplicity, suppose $\mathbf{O}_{i}$ is arranged in ascending order, i.e., $O_{ij} \leq O_{ik}$ if $ j\le k$. 
Given that the partial derivative of $L$ w.r.t. each $C_{ij}$ is a constant $O_{ij}$, the variation in $L$ remains monotonic when $C_{ij}$ is incremented or decremented. This indicates that $L$ reaches its minimum at the boundaries of its range. Consequently, the optimal solution of $\mathbf{C}_{i}$ is that only to have precisely one element set to 0 and the remainder set to 1. Notably $C_{il} = 0$ due to $O_{il}$ is the maximal element in $\mathbf{O}_{i}$. Therefore, the optimal solution for the optimization problem in Eq. (\ref{eq therotical tr question ij}) adheres to a zero-hot strategy, with the zero precisely being aligned with the peak value in $\mathbf{O}_i$.

\subsubsection{What properties does the collaborative term possess?}

The collaborative term is ingeniously crafted to establish a connection between the base and partner classifiers, manifesting as a trace term in its practical application. Its primary objective is to require $\mathbf{O}_i^\mathsf{T}\mathbf{C}_i=0$ for each sample $\mathbf{x}_i$. Generally, the relative value of $O_{ij}$ and $C_{ij}$ should be opposite, i.e., if $O_{ij}$ is large, $C_{ij}$ should be small. However, the collaborative term permits $C_{ij}$ to be small or even zero if $O_{ij}$ is small, which means that it works in a wild way. This gentle design is intentional: our goal is for the partner classifier to be aware of the base classifier's predictions without being overly swayed by them. Besides, this design also augments the classifier's resilience against faults. To circumvent $\mathbf{C}_{i}$ to be triviality $\mathbf{0}_l$, we introduce the constraint of $\sum_{j} C_{ij}=l-1$.

Furthermore, upon deriving the optimal solution for the problem in Eq. (\ref{eq therotical tr question}) determined, we can further elucidate the collaborative term in PLCP. Specifically, as $\gamma$ in Eq. (\ref{eq: partner classifier}) approaches positive infinity, each row of $\mathbf{C}$ is a strictly zero-hot vector, aligning the predicted ground-truth labels for each training sample with those of the base classifier, for the zero positions of $\mathbf{C}$ are the same as the maximum value positions in $\mathbf{O}$. However, given that $\gamma$ is only a small positive value distinctly short of infinity, the elements of $\mathbf{C}$ are values between 0 and 1 rather than being binary.

Therefore, we can conclude that the collaborative term operates as a wild agent, facilitating a soft and moderated interaction between the classifiers.

\subsubsection{Should the collaborative term be aggressive?}

As analysed above, the collaborative term links the two classifiers in a wild manner. Nonetheless, this raises a pertinent question: would a more aggressive collaborative term enhance performance?

To explore the impact of an aggressive collaborative term on performance, we consider the methodology outlined in \cite{jia2023complementary}, which links two classifiers through an aggressive term, denoted as $\|\mathbf{O}+\mathbf{C}-\mathbf{1}_{n\times l}\|_F^2$. This aggressive term significantly diverges from our collaborative term. It compels $\mathbf{O}_i + \mathbf{C}_i=\mathbf{1}_l$ for a sample $\mathbf{x}_i$, indicating that if $\mathbf{O}_i$ is small, $\mathbf{C}_i$ should be large. We compare the performances with these two kinds of linking terms, and the results are shown in Table \ref{tab:aggressive}. It becomes apparent that employing the aggressive term within PLCP markedly detracts from performance compared to our collaborative term. Consequently, this analysis suggests that a collaborative term designed to facilitate ``wild'' rather than ``aggressive'' interactions yields superior results.

\subsection{Analysis of the Blurring Mechanism \label{sec: blur ana}}
\subsubsection{Why can blurring mechanism blur?}

Suppose $a$ and $b$ are labeling confidences of two candidate labels of a sample. The difference between $a$ and $b$ after the blurring mechanism is
\begin{equation}
\frac{e^{e^ka}-e^{e^kb}}{e^{e^ka}+e^{e^kb}}= 1 - \frac{2}{e^{e^k (a-b)}+1},
\end{equation}
We want to prove the following inequality always holds when $k<0$:
\begin{equation}
    a-b > 1 -\frac{2}{e^{e^k (a-b)}+1}
    \label{eq: inequality}
\end{equation}
Define the function $\mathcal{F}(x)$ as $\mathcal{F}(x) = x + \frac{2}{e^{e^k x}+1}-1$ where $x=a-b>0$, if $k<0$, we have
\begin{equation}
    \frac{d\mathcal{F}}{dx} = 1 - \frac{e^k}{2\cosh^2(\frac{e^kx}{2})}>1 - \frac{1}{2} >0,
\end{equation}
which means that $\mathcal{F}(x)$ is monotonically increasing when $x>0$. Therefore, $\mathcal{F}(x)>\mathcal{F}(0)=0$, i.e., we have shown that the inequality in (\ref{eq: inequality}) is true when $k<0$. Q.E.D.

\subsubsection{Why do we need the blurring mechanism?}

As analysed in Sec. \ref{sec: colla ana}, significant discrepancies between elements $a$ and $b$ in $\mathbf{O}$ lead to significant differences between the corresponding elements $\hat{a}$ and $\hat{b}$ in $\mathbf{C}$, often inverting their relative magnitudes (i.e., if $a>b$, then typically $\hat{a}<\hat{b}$). Ideally, the partner classifier should learn non-candidate label information while considering the output of the base classifier, allowing it to supplement and improve its performance.
However, if we directly pass the predictions of the base classifier to the collaborative term without blurring, the labeling confidences of different labels predicted by the base classifier may differ greatly, causing the partner classifier to be greatly influenced by the base classifier. In extreme cases, this could result in the partner classifier mirroring the base classifier’s label predictions exactly.

To mitigate this, we need to design a blurring mechanism that can blur the outputs, so that elements with significant differences become closer after blurring. This blurring mechanism adjusts the absolute values of elements while preserving their relative order, ensuring the base classifier's influence exists but influence the partner classifier in a gentle way. Therefore, the partner classifier can utilize the non-candidate information to better assist in disambiguation and appeal, rather than dominantly guided by the base classifier’s outputs.

\begin{table*}[ht]\footnotesize
    \centering
    \caption{Two representative errors in SURE and LALO. ``$\rightarrow$'' means continuously increasing or decreasing.}
    \begin{tabular}{ccccc}
    \hline\hline
         Confidence & False Positive Label (FP) & Ground-truth Label (GT) & FP coupled with PLCP &	GT coupled with PLCP \\\hline
         SURE & 0.5000 $\rightarrow$ 0.5334 & 0.5000 $\rightarrow$ 0.4666 & 0.5334 $\rightarrow$ 0.4177 & 0.4666 $\rightarrow$ 0.5823 \\
         SURE & 0.6127 $\rightarrow$ 0.5327 & 0.3873 $\rightarrow$ 0.4673 & 0.5327 $\rightarrow$ 0.4881 & 0.4673 $\rightarrow$ 0.5119 \\
         LALO & 0.2500 $\rightarrow$ 0.3241 & 0.2500 $\rightarrow$ 0.2289 & 0.3241 $\rightarrow$ 0.2472 & 0.2289 $\rightarrow$ 0.3227\\
         LALO & 0.5625 $\rightarrow$ 0.5081 & 0.3181 $\rightarrow$ 0.4362 & 0.5081 $\rightarrow$ 0.4474 & 0.4362 $\rightarrow$ 0.4781 \\\hline\hline
    \end{tabular}
    \label{tab:representative error}
\end{table*}

\subsection{More Details of the Representative Errors}
Generally, the labeling confidence is initialized following Eq. (\ref{eq confidence initialize}), employing an averaging strategy. Nevertheless, as illustrated in Fig. \ref{fig:false positive}, the initial confidence levels of the two labels are unequal. This difference arises because PL-AGGD bases the initialization of labeling confidences on the local geometric structure among instances, resulting in a non-uniform strategy to set initial confidence levels.

Furthermore, the two representative errors highlighted in the introduction are not exclusive to PL-AGGD. Although PL-AGGD serves as a case study for these errors, they are prevalent across other PLL methods such as SURE and LALO, which also exhibit vulnerability to these issues. To substantiate this point, Table. \ref{tab:representative error} details how the confidence levels of labels for certain mislabeled samples in the data set Lost change under different confidence initialization strategies (i.e., average or non-uniform). Notably, our observation unveils that both SURE and LALO are susceptible to these two representative errors, indicating that these challenges are not isolated to a single method but are common across various PLL approaches. This observation emphasizes the universal challenge PLL methods face in identifying and correcting mislabeled samples, highlighting a fundamental obstacle in PLL.
\section{Related Work}\label{sec:related}

\subsection{Partial Label Learning}

Partial label learning (PLL), also known as superset-label learning \cite{liu2012conditional,liu2014learnability} or ambiguous label learning \cite{hullermeier2006learningPL-KNN, zeng2013learningSoccerplayer}, is a representative weakly supervised learning framework which learns from inaccurate supervision information. In partial label learning, each instance is associated with a set of candidate labels with only one being ground-truth and others being false positive. As the ground-truth label of a sample conceals in the corresponding candidate label set, which can not be directly acquired during the training process, partial label learning task is a quite challenging problem.

To tackle the mentioned challenge, existing works mainly focus on disambiguation \cite{feng2019partialsssfagdsg, nguyen2008classificationplsvm, 2015Solvingipal, wang2021adaptivePLAGGD, fan2021partial, xu2019partial,zhang2022disambiguation,qian2023disambiguation}, which can be broadly divided into two categories: averaging-based approaches and identification-based approaches. For the averaging-based approaches \cite{hullermeier2006learningPL-KNN, cour2011learning, 2015Solvingipal}, each candidate label of a training sample is treated equally as the ground-truth one and the final prediction is yielded by averaging the modeling outputs. For instance, PL-KNN \cite{hullermeier2006learningPL-KNN} averages the candidate labels of neighboring samples to make the prediction. This kind of approach is intuitive, however, it can be easily influenced by false positive candidate labels which results in inferior performance. For identification-based approaches, \cite{2018Leveraginglalo, feng2019partialsssfagdsg, nguyen2008classificationplsvm, jin2002learninggggggg, yu2016maximumhhhh}, the ground-truth label is treated as a latent variable and can be identified through an iterative optimization procedure such as EM. Moreover, labeling confidence based strategy is proposed in many state-of-the-art identification based approaches for better disambiguation. \cite{2015Solvingipal} and \cite{wang2021adaptivePLAGGD} construct a similarity graph based on the feature space to generate labeling confidence of candidate labels. 

Recently, models based on deep learning have been increasingly employed for disambiguation for PLL tasks \cite{lv2020progressiveproden, xu2021instancevalen, he2022partialplls,wu2022revisiting, lyu2022deep,xia2023towards}. PICO \cite{wang2022pico} is a contrastive learning-based approach devised to tackle label ambiguity in partial label learning. This method seeks to discern the valid label from the candidate set by utilizing contrastively learned embedding prototypes. Lv et al. proposed PRODEN \cite{lv2020progressiveproden}, where the simultaneous updating of the model and identification of true labels are seamlessly integrated. Furthermore, He et al. introduced a partial label learning method based on semantic label representations in \cite{he2022partialplls}. This method employs a novel weighted calibration rank loss to facilitate label disambiguation. By leveraging label confidence, the approach weights the similarity towards all candidate labels and subsequently yields a higher similarity of candidate labels in comparison to each non-candidate label.

\vspace{-3mm}
\subsection{Relations to CLL}

The concept of leveraging non-candidate labels, also known as complementary labels, is not new in the field of machine learning. Previous research on complementary-label learning (CLL) \cite{ishida2017learning, ishida2019complementary,gao2021discriminative, katsura2020bridging} primarily focused on scenarios involving a single complementary label. In contrast to CLL, PLL entails generating complementary labels from the set of non-candidate labels, serving the purpose of appeal and disambiguation in PLL.

The research focus on Complementary Labels (CL) in CLL \cite{ishida2017learning, ishida2019complementary,gao2021discriminative, katsura2020bridging} is primarily motivated by the inherent challenge it presents. Operating within a single CL framework allows for data efficiency but introduces significant complexity. In theory, CLL strategies could be modified for PLL applications. It's crucial, however, to underline the enduring challenges of PLL, even when employing a limited set of labels as candidates. A classifier trained under conditions where only one label is deemed ``invalid'' while the others are considered ``candidates'' is prone to sub-optimal performance. Therefore, although CLL methodologies might be theoretically applicable to PLL situations, their practical efficacy in such contexts might be constrained.

\vspace{-3mm}

\subsection{Relations to MCLL}
Notably, a recent setting known as Multiple Complementary Label Learning (MCLL) \cite{feng2020learning} has emerged. MCLL introduces the capability to apply the negative learning loss \cite{kim2019nlnl} to each CL associated with the same instance. This loss aims to minimize the class-posterior probability of each CL, effectively driving these probabilities towards zero. It's worth noting that in the original formulation of PLL \cite{jin2002learning}, the primary goal was to maximize the cumulative probabilities of all potential candidate labels for each instance towards one, while leaving the precise distribution among these candidates (be it uniform or skewed towards a singular "winner") unspecified. In this sense, MCLL and the original PLL share the exact same problem setting with the same objective.

However, it's essential to recognize that the implemented algorithms may introduce more specific training objectives beyond the default goal, such as identifying the valid labels. This leads to PLL and MCLL being distinct research topics, each with its unique algorithms and nuances.

\vspace{-2mm}

\section{Conclusion}\label{sec:con}
In this paper, we propose the first appeal-based framework PLCP, which aims to identify and rectify mislabeled samples for the existing PLL classifiers. Specifically, a partner classifier is introduced and a novel collaborative term is designed to link the base classifier and the partner classifier, which enables mutual supervision between the two classifiers. A blurring mechanism is involved in this paradigm for better disambiguation. Comprehensive experiments validate the outstanding performance of PLCP coupling with stand-alone approaches and deep-learning based methods, which further validates that the mislabeled examples can be identified and corrected by PLCP. Further analyses are provided for better understanding of this framework. In the future, it is also interesting to investigate other appeal-based methods to identify and correct mislabeled samples.

\vspace{-3mm}

\bibliographystyle{IEEEtran}
\bibliography{main}

\end{document}

\title{Disambiguation and Appeal: \\ 
Addressing Ambiguity and Error Correction in Partial Label Learning\\ Supplementary Material}

\author{Chongjie Si, Zekun Jiang, Xuehui Wang, Yan Wang, Xiaokang Yang,~\IEEEmembership{~Fellow,~IEEE}, Yuheng Jia,~\IEEEmembership{~Member,~IEEE}, Min-ling Zhang, ~\IEEEmembership{Senior~Member,~IEEE}, Wei Shen
        % <-this % stops a space
\IEEEcompsocitemizethanks{
\IEEEcompsocthanksitem C. Si, Z. Jiang, X. Wang, X. Yang, W. Shen are with MoE Key Lab of Artificial Intelligence, AI Institute, Shanghai Jiao Tong University, Shanghai 200240, China. \protect E-mail: \{chongjiesi, zkjiangzekun.cmu, wangxuehui, xkyang, wei.shen\}@sjtu.edu.cn.
\IEEEcompsocthanksitem Y. Wang is with the Shanghai Key Lab of Multidimensional Information Processing, East China Normal University, Shanghai 200241, China. \protect E-mail: ywang@cee.ecnu.edu.cn.
\IEEEcompsocthanksitem Y. Jia and M. Zhang are with the School of Computer Science and Engineering, Southeast University, Nanjing 210096, China. \protect E-mail: \{yhjia, zhangml\}@seu.edu.cn.
}
}

% The paper headers
\markboth{Journal of \LaTeX\ Class Files,~Vol.~14, No.~8, August~2021}%
{Shell \MakeLowercase{\textit{et al.}}: A Sample Article Using IEEEtran.cls for IEEE Journals}

%\IEEEpubid{0000--0000/00\$00.00~\copyright~2021 IEEE}
% Remember, if you use this you must call \IEEEpubidadjcol in the second
% column for its text to clear the IEEEpubid mark.

\section{Introduction}

\begin{figure}
    \centering
    \includegraphics[scale=0.15]{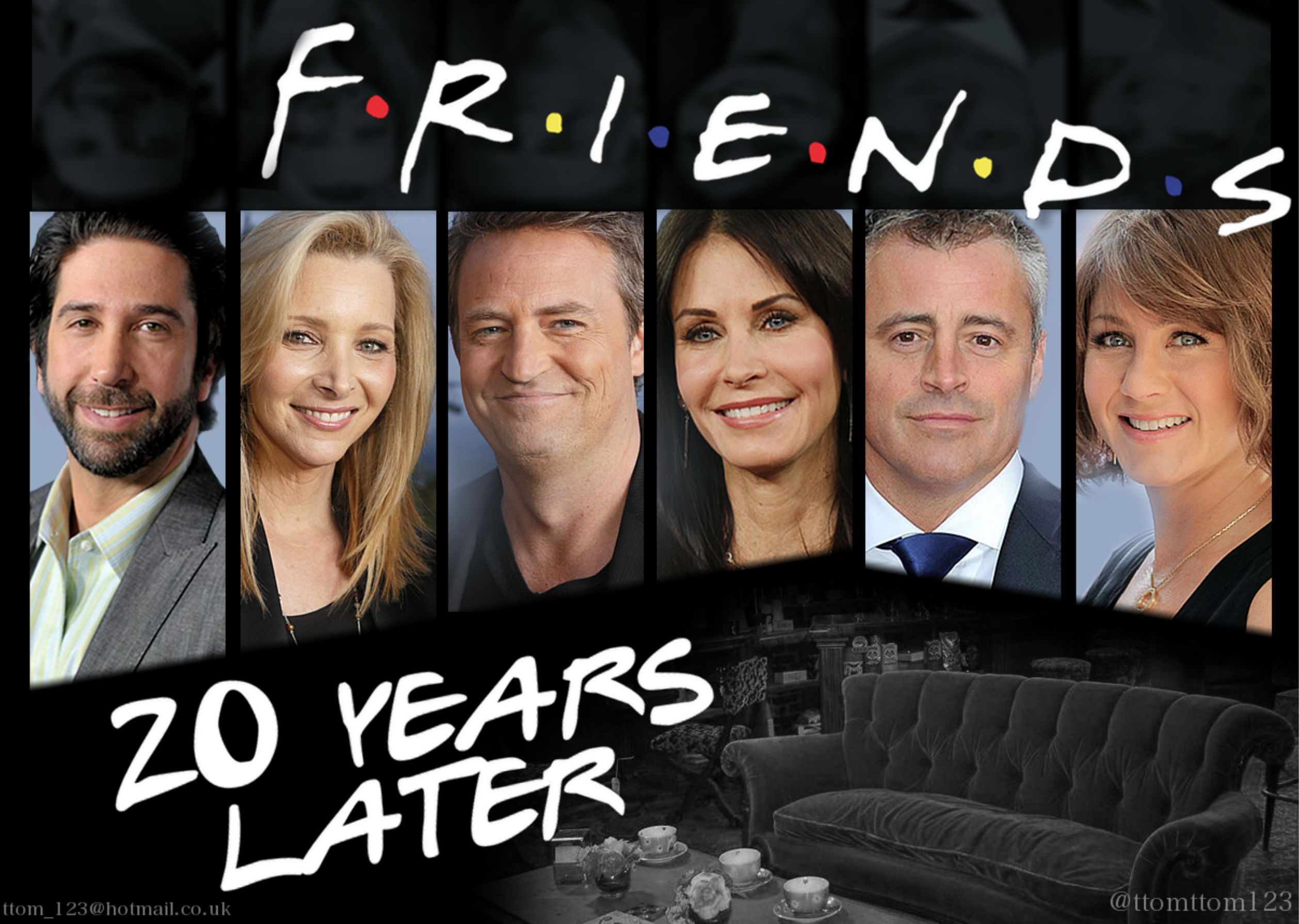}
    \caption{Partial label learning in the automatic face naming task. The faces in this image can be automatically detected by a face detector, and then each face is assigned with the candidate name label set extracted from the script: Jennifer Aniston, Courteney Cox, Lisa Kudrow, Matthew Perry, David Schwimmer, and Matt LeBlanc, where only one is ground-truth.}
    \label{fig:example PLL}
\end{figure}

In 2002, Jin and Ghahramani \cite{jin2002learning} pioneered the concept of Partial Label Learning (PLL), a groundbreaking framework in which each instance is associated with multiple candidate labels, among which only one represents the actual ground-truth. Over the past two decades, the field of PLL has experienced remarkable growth \cite{cour2011learning,han2018co22,jin2002learning,papandreou2015weaklysfs,ren2018learning,zhou2018brief,zhu2009introductionddd,chai2019large2462, li2019towards, li2019safe}, driven by the increasing necessity to accurately discern the valid label from a set of potential candidates in diverse real-world applications. A notable example, as shown in Fig. \ref{fig:example PLL}, 
is the automatic face naming task \cite{zeng2013learningSoccerplayer,guillaumin2010multipleYahoonews}, where each facial image extracted from various media is linked to a list of names derived from corresponding titles or captions \cite{gong2022partialicml}. Another salient application is facial age estimation: for each human face, the ages annotated by crowd-sourcing labelers are considered as candidate labels \cite{panis2016overviewFg-net}.

Evidently, the principal challenge in PLL lies in the inherent obscurity of the ground-truth label, which conceals within the candidate labels and remains inaccessible during the training phase. 
To surmount this obstacle, the most extensively researched and pivotal strategy in PLL-\textit{\textbf{Disambiguation}}-emerges as critical. Existing methods predominantly achieve disambiguation by differentiating the labeling confidences of each candidate label to identify the ground truth. This process typically relies on an alternative and iterative optimization algorithm for updating the classifier's parameters. For instance, LALO \cite{2018Leveraginglalo} implements constrained local consistency to differentiate the candidate labels, PL-AGGD \cite{wang2021adaptivePLAGGD} employs a similarity graph for effective disambiguation and PL-CL \cite{jia2023complementary} adopts complementary information to help disambiguation. However, a significant yet rarely studied question comes to the front: \textit{can the instances that are incorrectly disambiguated, i.e., mislabeled, have the opportunity to be rectified?} More specifically, \textit{can a classifier correct a false positive candidate label (i.e., invalid candidate label) with a large or upward-trending labeling confidence at a later stage?} To explore this question, we conduct experiments on a real-world data set Lost \cite{cour2009learningLost} and record the labeling confidences of several candidate labels generated by PL-AGGD \cite{wang2021adaptivePLAGGD} in each iteration, which is shown in Fig. \ref{fig:false positive}. Our findings reveal some intriguing phenomena: 
\begin{itemize}
    \item Each candidate label's labeling confidence is likely to continually increase or decrease until convergence.
    \item For a false positive candidate label with a large labeling confidence, although its confidence may decrease in subsequent iterations, the confidence remains substantial and can easily lead to the incorrect identification of the ground truth label.
\end{itemize}

\begin{figure*}
  \centering
  \subfloat[]{
		\includegraphics[scale=0.17]{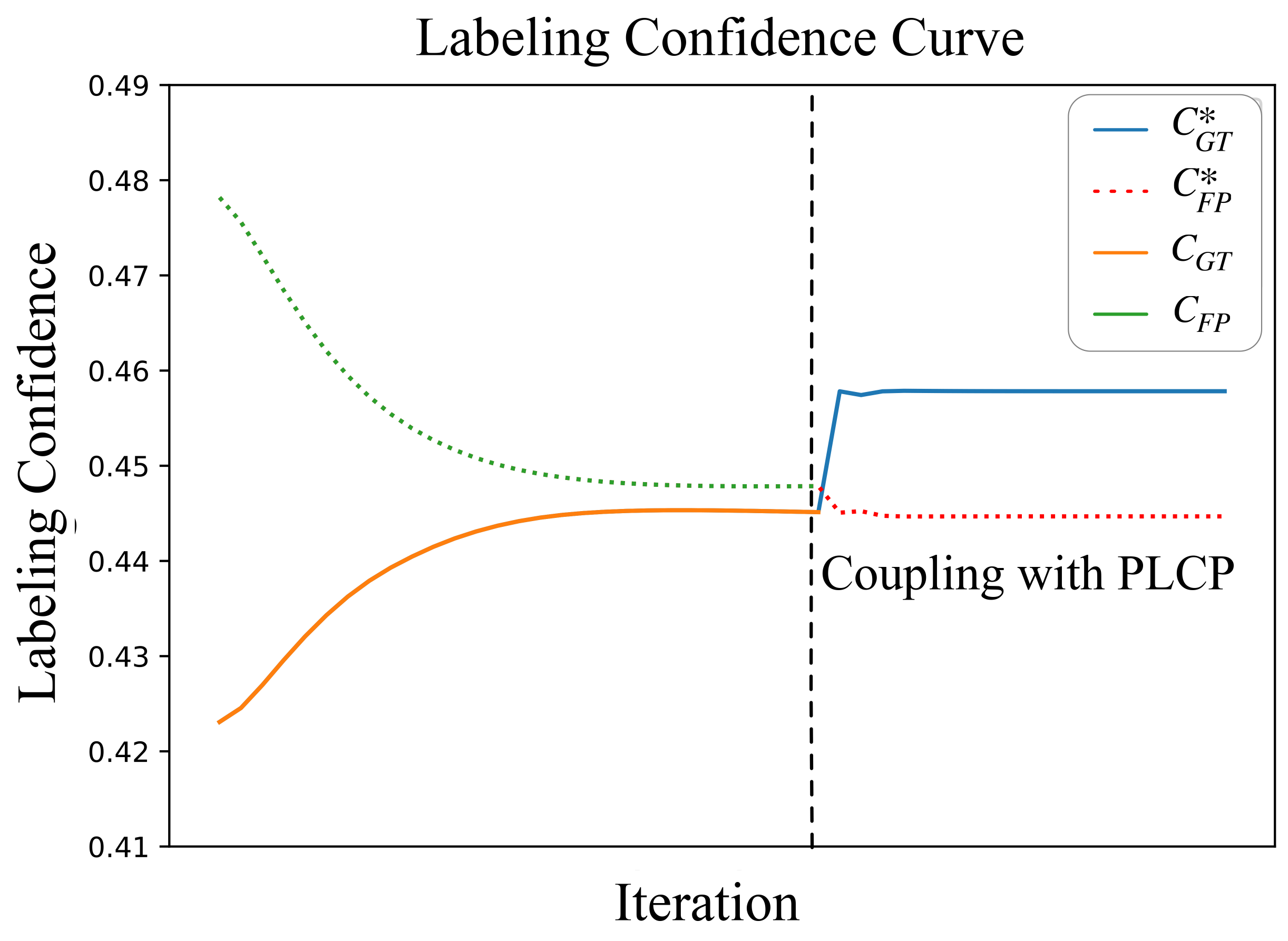}}
    \subfloat[]{
		\includegraphics[scale=0.17]{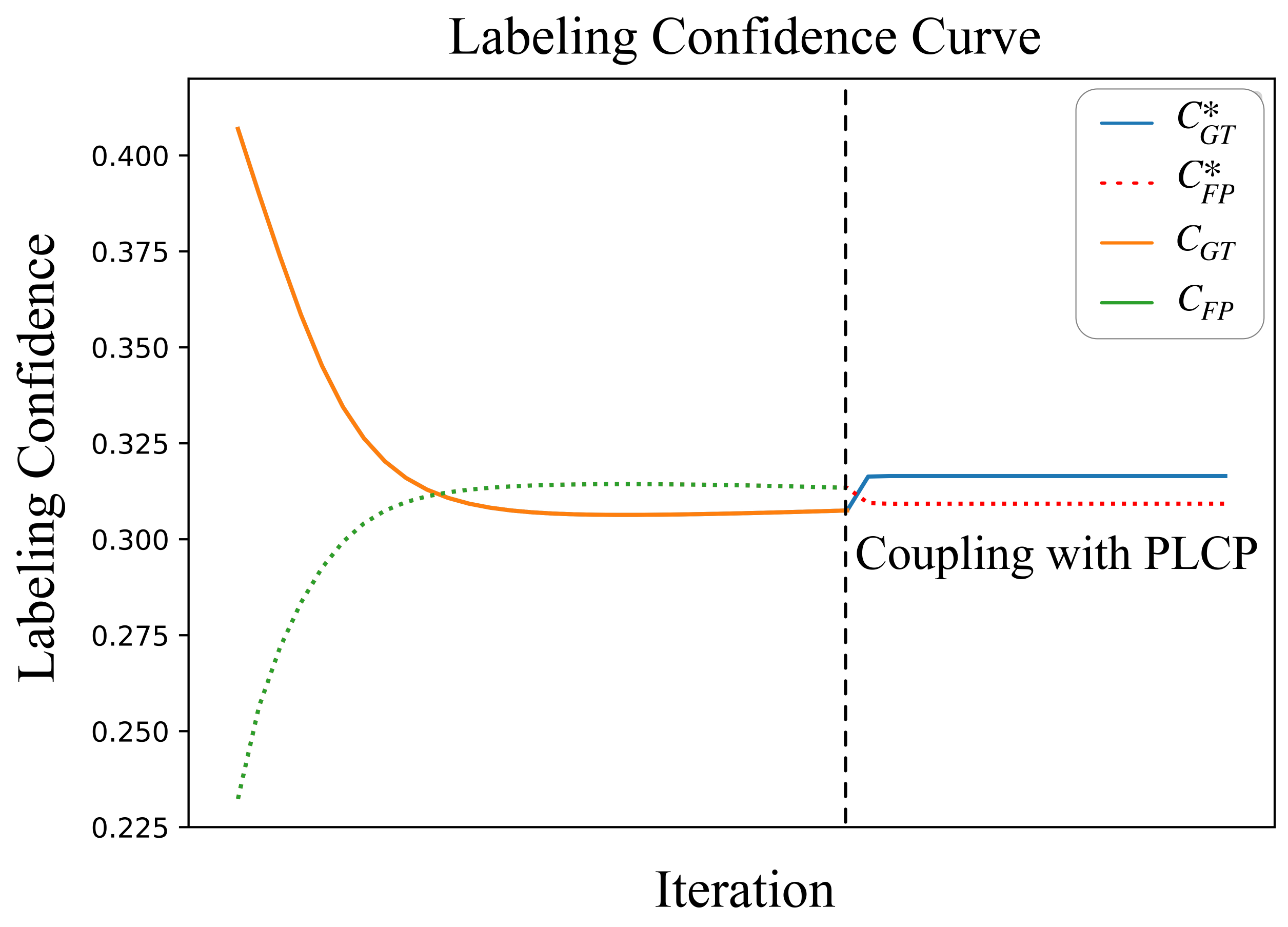}}
  \caption{Two representative errors a typical PLL classifier, e.g., PL-AGGD \cite{wang2021adaptivePLAGGD} may make. $C_{GT}$ and $C_{GT}^*$ stand for the labeling confidence of the ground-truth label generated by PL-AGGD and PL-AGGD coupled with PLCP, and $C_{FP}$ and $C_{FP}^*$ stand for that of a false positive label predicted by PL-AGGD and PL-AGGD coupled with PLCP. (a). For a false positive candidate label with a large labeling confidence, although its confidence may decrease properly, it could still be larger than the ground-truth one's. (b). The labeling confidence of a false positive candidate label keeps increasing and becomes the largest, which misleads the final prediction. When coupled with PLCP, the labeling confidence of each candidate label generated by the partner classifier is adopted as the supervision to help PL-AGGD correct these errors, which results in a mutation in the figures.}
  \label{fig:false positive}
\end{figure*}

The observed phenomena suggest that once the labeling confidence of a false positive candidate label increases, it becomes difficult to decrease in the subsequent iterations. Furthermore, even if the confidence of a false positive candidate label decreases appropriately, it may still be recognized as the ground truth one, as its initial labeling confidence remains large and continues to be greater than the confidence of the ground truth label upon convergence. As a result, correcting mislabeled samples for a PLL classifier itself proves to be quite challenging.

We believe that these mislabeled samples should be afforded the opportunity to be corrected. In other words, these samples should have a chance to ``appeal'' on their behalf. In this case, the classifier's disambiguation ability can also be further enhanced. Consequently, in this paper, we introduce another major strategy in PLL: \textit{\textbf{Appeal}}, and propose the first appeal-based framework.

Specifically, our framework PLCP, referring to \textbf{P}artial \textbf{L}abel Learning with a \textbf{C}lassifier as  \textbf{P}artner, provides each instance with the chance to appeal. Given a classifier from any existed PLL approach as the base classifier, as shown in Fig. \ref{fig:framework_plcp}, PLCP integrates an additional partner classifier. This partner classifier assists the base classifier in identifying and rectifying mislabeled samples, offering more precise and complementary information to the base classifier, thereby enhancing disambiguation and fostering mutual supervision between the two classifiers. The design of the partner classifier is crucial in whether mislabeled samples could appeal for themselves, as the partner classifier's feedback significantly influences the base classifier's capability to recognize and amend mislabeled instances. Since the information in non-candidate labels, which indicates that a set of labels DO NOT belong to a sample, is typically more precise yet often overlooked by the majority of existing works, the partner classifier is devised as a complementary classifier. This classifier specifies the labels that should not be assigned to a sample, thereby complementing the base classifier. Additionally, a collaborative term is also designed to link the base classifier and the partner classifier. 

During mutual supervision, the labeling confidence is first updated based on the base classifier's modeling output.  A blurring mechanism is then applied to this updated labeling confidence, introducing uncertainty. This could potentially diminish the high confidence of certain false positive candidate labels or elevate the low confidence of the actual ground truth. This updated labeling confidence subsequently serves as the supervision information to interact with the partner classifier, whose final output, in turn, supervises the base classifier. The predictions of the two classifiers, while distinct, are inextricably linked, enhancing the disambiguation ability of this paradigm in two opposing ways. With this mutual supervision paradigm, the instances with disambiguation errors have a higher likelihood to appeal successfully.

The preliminary results of this work are presented in \cite{si2023partial}. In this paper, we refine the method in \cite{si2023partial} and propose a more robust version of PLCP. Besides, comprehensive theoretical analyses of  PLCP are provided in this paper.
The rest of the paper is organized as follows. We first introduce the appeal-based framework PLCP in Sec. \ref{sec: method}, and then present the experimental results and ablation study in Sec. \ref{sec: exp}. Further analysis is provided in Sec. \ref{sec:fur}. We then review some related works in PLL and other related fields in Sec. \ref{sec:related}. Finally, conclusion is given in Sec. \ref{sec:con}
\input{sec/2.related work}
\input{sec/3.method1}
\input{sec/4.method2}
\maketitle
\IEEEdisplaynontitleabstractindextext
\IEEEpeerreviewmaketitle
\setcounter{page}{1}

\section{Two Representative Errors}
In this section, we explain why it is difficult for a PLL classifier to rectify those samples that are incorrectly disambiguated.
We conduct experiments on a real-world data set Lost \cite{cour2009learningLost} and record the labeling confidences of several candidate labels generated by PL-AGGD \cite{wang2021adaptivePLAGGD} in each iteration, which is shown in Fig. \ref{fig:false positive}. Our findings reveal some intriguing phenomena: 
\begin{itemize}
    \item Each candidate label's labeling confidence is likely to continually increase or decrease until convergence.
    \item For a false positive candidate label with a large labeling confidence, although its confidence may decrease in subsequent iterations, the confidence remains substantial and can easily lead to the incorrect identification of the ground truth label.
\end{itemize}

\begin{figure}
  \centering
  \subfloat{
		\includegraphics[scale=0.09]{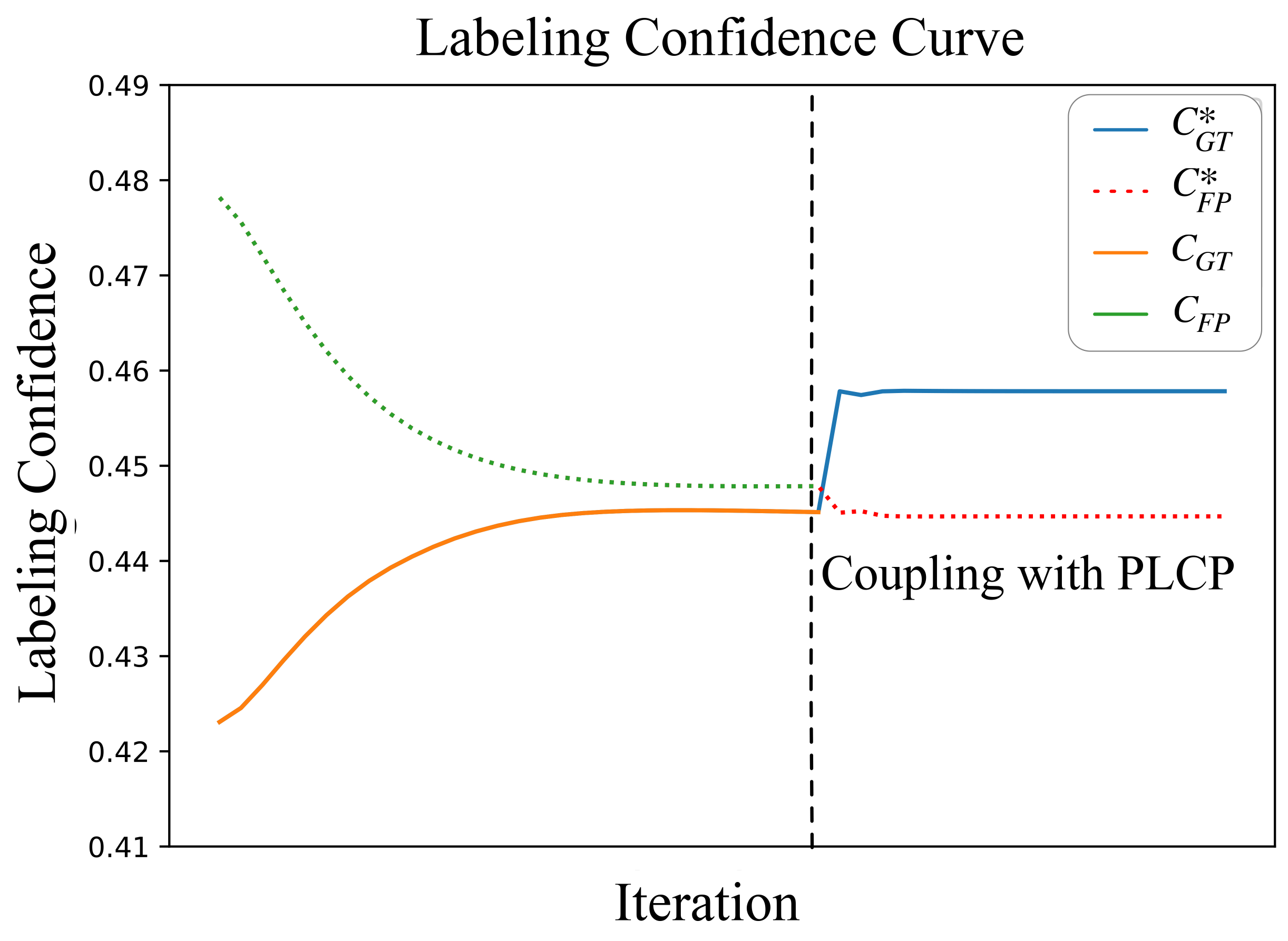}}
    \subfloat{
		\includegraphics[scale=0.09]{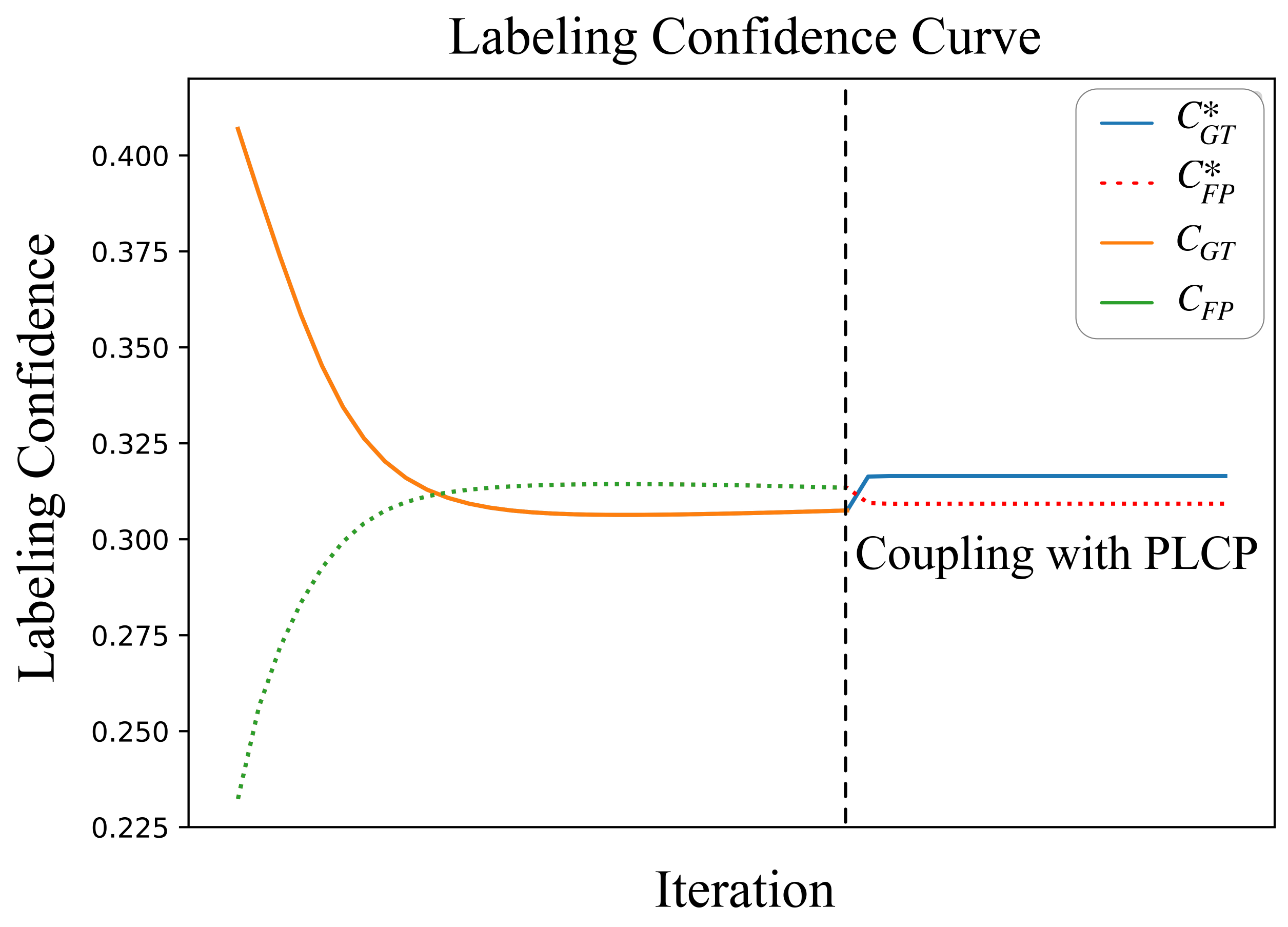}
  }
  \caption{Two representative errors a typical PLL classifier, e.g., PL-AGGD \cite{wang2021adaptivePLAGGD} may make. $C_{GT}$ and $C_{GT}^*$ stand for the labeling confidence of the ground-truth label generated by PL-AGGD and PL-AGGD coupled with PLCP, and $C_{FP}$ and $C_{FP}^*$ stand for that of a false positive label predicted by PL-AGGD and PL-AGGD coupled with PLCP. (a). For a false positive candidate label with a large labeling confidence, although its confidence may decrease properly, it could still be larger than the ground-truth one's. (b). The labeling confidence of a false positive candidate label keeps increasing and becomes the largest, which misleads the final prediction. When coupled with PLCP, the new labeling confidence of each candidate label generated by the partner classifier is adopted as the supervision to help PL-AGGD correct these errors, which results in a mutation in the figures.}
  \label{fig:false positive}
\end{figure}

The observed phenomena suggest that once the labeling confidence of a false positive candidate label increases, it becomes difficult to decrease in the subsequent iterations. Furthermore, even if the confidence of a false positive candidate label decreases appropriately, it may still be recognized as the ground truth one, as its initial labeling confidence remains large and continues to be greater than the confidence of the ground truth label upon convergence. As a result, correcting mislabeled samples for a PLL classifier itself proves to be quite challenging.

Besides, the two representative errors elucidated above do not only occur in PL-AGGD. While we employ PL-AGGD as an illustrative example to demonstrate these errors, they extend beyond this specific method. In reality, other PLL methods like SURE and LALO are also susceptible to these two representative errors. To substantiate this point, Tab. \ref{tab:representative error} outlines the alterations in the confidence of labels for certain mislabeled samples on data set Lost. Notably, our observation unveils that both SURE and LALO are prone to experiencing these two errors. Therefore, it is evident that these errors are not exclusive to any one specific method, but rather are shared across various PLL approaches. Moreover, this broader perspective underscores the inherent difficulty faced by PLL methods when it comes to identifying and rectifying mislabeled samples. 

To bridge this gap, PLCP is proposed, acting as a supportive framework for all kinds of PLL classifiers to effectively correct mislabeled samples.

\begin{table*}[ht]\footnotesize
    \centering
    \begin{tabular}{ccccc}
    \hline\hline
         Confidence & False Positive Label (FP) & Ground-truth Label (GT) & FP coupled with PLCP &	GT coupled with PLCP \\\hline
         SURE & 0.5000 $\rightarrow$ 0.5334 & 0.5000 $\rightarrow$ 0.4666 & 0.5334 $\rightarrow$ 0.4177 & 0.4666 $\rightarrow$ 0.5823 \\
         SURE & 0.6127 $\rightarrow$ 0.5327 & 0.3873 $\rightarrow$ 0.4673 & 0.5327 $\rightarrow$ 0.4881 & 0.4673 $\rightarrow$ 0.5119 \\
         LALO & 0.2500 $\rightarrow$ 0.3241 & 0.2500 $\rightarrow$ 0.2289 & 0.3241 $\rightarrow$ 0.2472 & 0.2289 $\rightarrow$ 0.3227\\
         LALO & 0.5625 $\rightarrow$ 0.5081 & 0.3181 $\rightarrow$ 0.4362 & 0.5081 $\rightarrow$ 0.4474 & 0.4362 $\rightarrow$ 0.4781 \\\hline\hline
    \end{tabular}
    \caption{Two representative errors in SURE and LALO. ``$\rightarrow$'' means continuously increasing or decreasing.}
    \label{tab:representative error}
\end{table*}

\section{Numerical Solution of PL-CL}
The optimization problem in Eq. (\ref{eq objective function}) has seven variables with different constraints, we therefore adopt an alternative and iterative manner to solve it.

\subsection{Update W, b and $\hat{\mathbf{W}}$, $\hat{\mathbf{b}}$}\label{sec kernel}

With other variables fixed, problem (\ref{eq objective function}) with respect to $\mathbf{W}$ and $\mathbf{b}$ can be reformulated as

\begin{equation}
    \min_{\mathbf{W}, \mathbf{b}}\quad   \left\|\mathbf{XW}+\mathbf{1}_n\mathbf{b}^\mathsf{T} - \mathbf{P} \right\|_F^2 + \lambda  \left\|\mathbf{W}\right\|_F^2, 
    \label{eq W b problem}
\end{equation}
which is a regularized least squares problem and the corresponding closed-form solution is 

\begin{equation}
    \begin{split}
        \mathbf{W} &= \left(\mathbf{X}^\mathsf{T}\mathbf{X}+\lambda\mathbf{I}_{n\times n} \right)^{-1}\mathbf{X}^\mathsf{T}\mathbf{P}\\
        \mathbf{b}&=\frac{1}{n}\left(\mathbf{P}^\mathsf{T}\mathbf{1}_n - \mathbf{W}^\mathsf{T}\mathbf{X}^\mathsf{T}\mathbf{1}_n \right),
    \end{split}
\end{equation}
where $\mathbf{I}_{n\times n}$ is an identity matrix with the size of $n\times n$. Similarly, the problem (\ref{eq objective function}) with respect to $\hat{\mathbf{W}}$ and $\hat{\mathbf{b}}$ is

\begin{equation}
    \min_{\hat{\mathbf{W}}, \hat{\mathbf{b}}}\quad  \|\mathbf{X}\hat{\mathbf{W}} + \mathbf{1}_n\hat{\mathbf{b}}^{\mathsf{T}} - \hat{\mathbf{P}}\|_F^2 + \frac{\lambda}{\alpha_1} \|\hat{\mathbf{W}}\|_F^2,
\end{equation}
which has a similar analytical solution as Eq. (\ref{eq W b problem}), i.e.,

\begin{equation}
    \begin{split}
        \hat{\mathbf{W}} &= \left(\mathbf{X}^\mathsf{T}\mathbf{X}+\frac{\lambda}{\alpha_1}\mathbf{I}_{n\times n}\right)^{-1}\mathbf{X}^\mathsf{T}\hat{\mathbf{P}}\\
        \hat{\mathbf{b}}&=\frac{1}{n}\left(\hat{\mathbf{P}}^\mathsf{T}\mathbf{1}_n - \hat{\mathbf{W}}^\mathsf{T}\mathbf{X}^\mathsf{T}\mathbf{1}_n\right).
    \end{split}
\end{equation}

\noindent
\textbf{Kernel Extension} The above linear model may be too simple to tackle the complex relationships between the instances to labels, therefore we extend the linear model to a kernel version. Suppose $\phi(\cdot):\mathbb{R}^q \rightarrow \mathbb{R}^h$ represents a feature transformation that maps the feature space to a higher dimensional space, and $\mathbf{\Phi}= [\phi(\mathbf{x}_1), \phi(\mathbf{x}_2),...,\phi(\mathbf{x}_n) ]^\mathsf{T}$ denotes the feature matrix in the higher dimensional space. With the feature mapping, we can rewrite problem (\ref{eq W b problem}) as follows:

\begin{equation}
\begin{split}
    \min_{\mathbf{W}, \mathbf{b}}\quad& \|\mathbf{M} \|_F^2 + \lambda \|\mathbf{W}\|_F^2\\
    {\rm s.t.}\quad& \mathbf{M} = \mathbf{\Phi}\mathbf{W}+\mathbf{1}_n\mathbf{b}^\mathsf{T} - \mathbf{P},
    \end{split}
    \label{eq A b problem}
\end{equation}
where $\mathbf{M} = [\mathbf{m}_1,\mathbf{m}_2,...,\mathbf{m}_n ] \in \mathbb{R}^{n \times l}$. The Lagrangian function of Eq. (\ref{eq A b problem}) is formulated as 

\begin{equation}
\begin{split}
    \mathcal{L}(\mathbf{W},\mathbf{b},\mathbf{M},\mathbf{A}) = \|\mathbf{M} \|_F^2 + \lambda \|\mathbf{W}\|_F^2 \\- {\rm tr}\left(\mathbf{A}^{\mathsf{T}}(\mathbf{\Phi}\mathbf{W}+\mathbf{1}_n\mathbf{b}^\mathsf{T} - \mathbf{P}-\mathbf{M})\right),
\end{split}
\end{equation}
where $\mathbf{A} \in \mathbb{R}^{n\times l}$ is the Lagrange multiplier. ${\rm tr}(\cdot)$ is the trace of a matrix. Based on the KKT conditions, we have

\begin{equation}
    \begin{split}
        &\frac{\partial \mathcal{L}}{\partial \mathbf{W}} = 2\lambda\mathbf{W} -\mathbf{\Phi}^{\mathsf{T}}\mathbf{A}=0, \frac{\partial \mathcal{L}}{\partial \mathbf{b}} = \mathbf{A}^{\mathsf{T}}\mathbf{1}_n=0 \\
       & \frac{\partial \mathcal{L}}{\partial \mathbf{M}} = 2\mathbf{M}+ \mathbf{A}=0,   \frac{\partial \mathcal{L}}{\partial \mathbf{A}} = \mathbf{\Phi}\mathbf{W}+\mathbf{1}_n\mathbf{b}^\mathsf{T} - \mathbf{P}-\mathbf{M}=0.
       \label{eq A b gradient}
    \end{split}
\end{equation}
Define a kernel matrix $\mathbf{K}=\mathbf{\Phi}\mathbf{\Phi}^{\mathsf{T}}$ with its element $k_{ij} = \phi(\mathbf{x}_i)\phi(\mathbf{x}_j) = \mathcal{K}(\mathbf{x}_i,\mathbf{x}_j)$, where $\mathcal{K}(\cdot,\cdot)$ is the kernel function. For PL-CL, we use Gaussian function $\mathcal{K}(\mathbf{x}_i,\mathbf{x}_j) = {\rm exp}(-\|\mathbf{x}_i - \mathbf{x}_j\|_2^2/(2\sigma^2))$ as the kernel function and set $\sigma$ to the average distance of all pairs of training instances. Then, we have

\begin{equation}
\begin{split}
    \mathbf{A} &= \left(\frac{1}{2\lambda} \mathbf{K} + \frac{1}{2}\mathbf{I}_{n\times n}\right)^{-1}\left(\mathbf{P} - \mathbf{1}_n\mathbf{b}^{\mathsf{T}}\right),\\
     \mathbf{b} &= \left(\frac{\mathbf{sP}}{\mathbf{s1}_n}\right)^{\mathsf{T}},
\end{split}
    \label{eq A b solution}
\end{equation}
where $\mathbf{s} =\mathbf{1}_n^{\mathsf{T}}\left(\frac{1}{2\lambda} \mathbf{K} + \frac{1}{2}\mathbf{I}_{n\times n}\right)^{-1}$. We can also obtain $\mathbf{W} = {\mathbf{\Phi}^{\mathsf{T}} \mathbf{A}}/{2\lambda}$ from Eq. (\ref{eq A b gradient}). The output of the model can be denoted by $\mathbf{H} = \mathbf{\Phi}\mathbf{W} +\mathbf{1}_n\mathbf{b}^{\mathsf{T}} = \frac{1}{2\lambda}\mathbf{KA} + \mathbf{1}_n\mathbf{b}^{\mathsf{T}}$.

As the optimization problem regarding $\hat{\mathbf{W}}$ and $\hat{\mathbf{b}}$ has the same form as problem (\ref{eq A b problem}), similarly, we have
\begin{equation}
\begin{split}
    \hat{\mathbf{A}} &= \left(\frac{\alpha_1}{2\lambda} \mathbf{K} + \frac{1}{2}\mathbf{I}_{n\times n}\right)^{-1}\left(\hat{\mathbf{P}} - \mathbf{1}_n\hat{\mathbf{b}}^{\mathsf{T}}\right),\\
    \hat{\mathbf{b}} &= \left(\frac{\hat{\mathbf{s}}\hat{\mathbf{P}}}{\hat{\mathbf{s}}\mathbf{1}_n}\right)^{\mathsf{T}}, 
\end{split}
\label{eq what bhat solution}
\end{equation}
where $\hat{\mathbf{s}} =\mathbf{1}_n^{\mathsf{T}}\left(\frac{\alpha_1}{2\lambda} \mathbf{K} + \frac{1}{2}\mathbf{I}_{n\times n}\right)^{-1}$, and the output of the complementary classifier is defined as $\hat{\mathbf{H}} = \frac{\alpha_1}{2\lambda}\mathbf{K}\hat{\mathbf{A}} + \mathbf{1}_n\hat{\mathbf{b}}^{\mathsf{T}}$.

\subsection{Update $\hat{\mathbf{P}}$}
Fixing other variables, the $\hat{\mathbf{P}}$-subproblem can be equivalently rewritten as

\begin{equation}
    \begin{split}
    \min_{\hat{\mathbf{P}}}\quad& \left\|\hat{\mathbf{P}} - \frac{\alpha_1\hat{\mathbf{H}} + \beta(\mathbf{1}_{n\times l}-\mathbf{P})}{\alpha_1+\alpha_2}\right\|_F^2 \\
    {\rm s.t.}\quad & \hat{\mathbf{Y}}\leq\hat{\mathbf{P}}\leq \mathbf{1}_{n\times l},
\end{split}
\label{eq Q problem reformat}
\end{equation}
where the solution is 

\begin{equation}
    \hat{\mathbf{P}} =  \mathcal{T}_1\left(\mathcal{T}_{\hat{\mathbf{Y}}}\left(\frac{\alpha_1\hat{\mathbf{H}} + \alpha_2(\mathbf{1}_{n\times l}-\mathbf{P})}{\alpha_1+\alpha_2}\right)\right).
    \label{eq Q solution}
\end{equation}
$\mathcal{T}_1$, $\mathcal{T}_{\hat{\mathbf{Y}}}$ are two thresholding operators in element-wise, i.e., $\mathcal{T}_1(m):=\min\{1,m\}$ with $m$ being a scalar and $\mathcal{T}_{\hat{\mathbf{Y}}}(m):=\max\{\hat{{y}}_{ij}, m\}$.

\subsection{Update G}

Fixing other variables, the $\mathbf{G}$-subproblem is rewritten as 

\begin{equation}
\begin{split}
    \min_{\mathbf{G}} \quad & \alpha_3\left\|\mathbf{X}^\mathsf{T}-\mathbf{X}^\mathsf{T}\mathbf{G}\right\|_F^2 + \alpha_4\left\|\mathbf{P}^\mathsf{T}-\mathbf{P}^\mathsf{T}\mathbf{G}\right\|_F^2\\
    {\rm s.t.}\quad & \mathbf{G}^\mathsf{T}\mathbf{1}_n = \mathbf{1}_n, \mathbf{0}_{n\times n} \leq \mathbf{G} \leq \mathbf{U}.
\end{split}
\label{eq G problem}
\end{equation}
Notice that each column of $\mathbf{G}$ is independent to other columns, therefore we can solve the $\mathbf{G}$-subproblem column by column. To solve the $i$-th column of $\mathbf{G}$, we have

\begin{equation}
\begin{split}
    \min_{\mathbf{G_{\cdot i}}} \quad & \alpha_3\left\|\mathbf{x}_i - \sum_{k_{ji}=1} g_{ji}\mathbf{x}_j \right\|_F^2 + \alpha_4\left\|\mathbf{p}_i-\sum_{k_{ji}=1} g_{ji}\mathbf{p}_j\right\|_F^2\\
    {\rm s.t.}\quad & \mathbf{G}_{\cdot i}^\mathsf{T}\mathbf{1}_n = 1, \mathbf{0}_{n} \leq \mathbf{G}_{\cdot i} \leq \mathbf{U}_{\cdot i}.
\end{split}
\label{eq G subproblem}
\end{equation}
As there are only $k$ non-zero elements in $\mathbf{G}_{\cdot i}$ that are supposed to be updated, which corresponds to the reconstruction coefficients of $\mathbf{x}_i$ by its top-$k$ neighbors, let $\hat{\mathbf{g}}_i\in \mathbb{R}^k$ denote the vector of these elements. Let $\mathcal{N}_i$ be the set of the indexes of these neighbors corresponding to $\hat{\mathbf{g}}_i$. Define matrix $\mathbf{D}^{x_i} = [\mathbf{x}_i-\mathbf{x}_{\mathcal{N}_{i(1)}}, \mathbf{x}_i-\mathbf{x}_{\mathcal{N}_{i(2)}},...,\mathbf{x}_i-\mathbf{x}_{\mathcal{N}_{i(k)}}]^\mathsf{T} \in \mathbb{R}^{k\times l}$ and $\mathbf{D}^{p_i} = [\mathbf{p}_i-\mathbf{p}_{\mathcal{N}_{i(1)}}, \mathbf{p}_i-\mathbf{p}_{\mathcal{N}_{i(2)}},...,\mathbf{p}_i-\mathbf{p}_{\mathcal{N}_{i(k)}}]^\mathsf{T} \in \mathbb{R}^{k\times l}$, let Gram matrices $\mathbf{B}^{x_i} = \mathbf{D}^{x_i}(\mathbf{D}^{x_i})^\mathsf{T} \in \mathbb{R}^{k\times k}$ and $\mathbf{B}^{p_i} = \mathbf{D}^{p_i}(\mathbf{D}^{p_i})^\mathsf{T} \in \mathbb{R}^{k\times k}$, we can transform the problem in Eq. (\ref{eq G subproblem}) into the following form:

\begin{equation}
    \begin{split}
    \min_{\hat{\mathbf{g}}_i} \quad & \hat{\mathbf{g}}_i^\mathsf{T}\left(\alpha_3 \mathbf{B}^{x_i} + \alpha_4 \mathbf{B}^{p_i} \right)\hat{\mathbf{g}}_i\\
    {\rm s.t.}\quad & \hat{\mathbf{g}}_i^\mathsf{T}\mathbf{1}_k = 1, \mathbf{0}_{k} \leq \hat{\mathbf{g}}_i \leq \mathbf{1}_k.
    \end{split}
    \label{eq G solution}
\end{equation}
The optimization problem in Eq. (\ref{eq G solution}) is a standard Quadratic Programming (QP) problem, and can be solved by off-the-shelf QP tools. $\mathbf{G}$ is updated by concatenating all the solved $\hat{\mathbf{g}}_i$ together.

\subsection{Update P}
With other variables fixed, the $\mathbf{P}$-subproblem is 

\begin{equation}
    \begin{split}
    \min_{{\mathbf{P}}} \quad & \left\|\mathbf{H}-\mathbf{P}\right\|_F^2 + \alpha_2\left\|\mathbf{E}-\hat{\mathbf{P}}-\mathbf{P}\right\|_F^2+ \alpha_4\left\|\mathbf{P}^\mathsf{T}-\mathbf{P}^\mathsf{T}\mathbf{G}\right\|_F^2\\
    {\rm s.t.}\quad & \mathbf{P}\mathbf{1}_q = \mathbf{1}_n, \mathbf{0}_{n\times l} \leq \mathbf{P} \leq \mathbf{Y},
    \end{split}
    \label{eq P problem}
\end{equation}
and we rewrite the problem in Eq. (\ref{eq P problem})
into the following form:

\begin{equation}
    \begin{split}
    \min_{{\mathbf{P}}} \quad & \left\|\mathbf{P}-\frac{\mathbf{H}+\alpha_2(\mathbf{E}-\mathbf{Q})}{1+\alpha_2}\right\|_F^2 + \frac{\alpha_4}{1+\alpha_2} \left \|\mathbf{P}^\mathsf{T}-\mathbf{P}^\mathsf{T}\mathbf{G} \right \|_F^2\\
    {\rm s.t.}\quad & \mathbf{P}\mathbf{1}_q = \mathbf{1}_n, \mathbf{0}_{n\times l} \leq \mathbf{P} \leq \mathbf{Y}.
    \end{split}
    \label{eq P rewritten problem}
\end{equation}
In order to solve the problem (\ref{eq P rewritten problem}), we denote $\widetilde{\mathbf{p}} = {\rm vec}\left(\mathbf{P}\right) \in [0,1]^{nl}$, where ${\rm vec}(\cdot)$ is the vectorization operator. Similarly, $\widetilde{\mathbf{o}} = {\rm vec}\left(\frac{\mathbf{H}+\beta(\mathbf{E}-\mathbf{Q})}{1+\beta} \right) \in \mathbb{R}^{nl}$ and $\widetilde{\mathbf{y}} = {\rm vec}(\mathbf{Y}) \in\{0,1\}^{nl}$. Let $\mathbf{T} = 2(\mathbf{I}_{n\times n}-\mathbf{G})(\mathbf{I}_{n\times n}-\mathbf{G})^\mathsf{T} \in \mathbb{R}^{n \times n}$ be a square matrix. Based on these notations, the optimization problem (\ref{eq P rewritten problem}) can be written as 

\begin{equation}
    \begin{split}
    \min_{\widetilde{\mathbf{p}}} \quad & \frac{1}{2} \widetilde{\mathbf{p}}^\mathsf{T} \left( \mathbf{C} + \frac{2(1+\alpha_2)}{\alpha_4}\mathbf{I}_{nl\times nl} \right)\widetilde{\mathbf{p}}-\frac{2(1+\alpha_2)}{\alpha_4}\widetilde{\mathbf{o}}^\mathsf{T}\widetilde{\mathbf{p}}\\
    {\rm s.t.}\quad & \sum_{i=1,i\% n = j, 0\leq j\leq n-1}^{nl}\widetilde{\mathbf{p}}_i=1, \mathbf{0}_{nl} \leq \widetilde{\mathbf{p}}\leq \widetilde{\mathbf{y}},
    \end{split}
    \label{eq P solution}
\end{equation}
where $\mathbf{C}\in \mathbb{R}^{nl\times nl}$ is defined as:

\begin{equation}
   {\mathbf{C}}=\left[\begin{array}{cccc}
{\mathbf{T}} & \mathbf{0}_{m \times m} & \cdots & \mathbf{0}_{m \times m} \\
\mathbf{0}_{m \times m} & {\mathbf{T}} & \ddots & \vdots \\
\vdots & \ddots & \ddots & \mathbf{0}_{m \times m} \\
\mathbf{0}_{m \times m} & \cdots & \mathbf{0}_{m \times m} & {\mathbf{T}}
\end{array}\right].
\end{equation}
As the problem in Eq. (\ref{eq P solution}) is a standard QP problem, it can be solved by off-the-shelf QP tools.

\subsection{The Overall Optimization Algorithm}
As a summary, PL-CL first initializes  $\mathbf{G}$. and $\mathbf{P}$ by solving problems (\ref{eq graph G}) and (\ref{eq graph P}) respectively, and initializes $\hat{\mathbf{P}}$ as:

\begin{equation}
\hat{{p}}_{i j}=\left\{\begin{array}{ll}
\frac{1}{l-\sum_{j} y_{i j}} & \text { if } \quad y_{i j}=0 \\
0 & \text { otherwise }.
\end{array}\right.
\label{eq Q ini}
\end{equation}
Then, PL-CL iteratively and alternatively updates one variable with the other fixed until the model converges.

For an unseen instance $\mathbf{x}^\ast$, the predicted label $y^\ast$ is 

\begin{equation}
    y^{*} = \mathop{\arg\max}_{k} \sum_{i=1}^n \frac{1}{2\lambda}\mathbf{A}_{ik}\mathcal{K}(\mathbf{x}^\ast, \mathbf{x}_i)+b_k.
    \label{eq prediction}
\end{equation}
The pseudo code of PL-CL is summarized in Algorithm \ref{alg:PLCL}.

\begin{algorithm}[tb]
   \caption{The pseudo code of PL-CL}
   \label{alg:PLCL}
\begin{algorithmic}
   \STATE {\bfseries Input:} Partial label data $\mathcal{D}$, hyper-parameter $\alpha$, $\beta$, $\mu$, $\gamma$ and $\lambda$, an unseen sample $\mathbf{x}^\ast$
   \STATE {\bfseries Output:} The predicted label $y^\ast$ of $\mathbf{x}^\ast$
   \STATE {\bfseries Process}:
   \STATE Initialize $\mathbf{G}$ according to Eq. (\ref{eq graph G}).
   \STATE Initialize $\mathbf{P}$ according to Eq. (\ref{eq graph P}).
   \STATE Initialize $\hat{\mathbf{P}}$ according to Eq. (\ref{eq Q ini}).
   \REPEAT
   \STATE Update $\mathbf{A}$, $\mathbf{b}$ according to Eq. (\ref{eq A b solution}).
   \STATE Update $\hat{\mathbf{A}}$, $\hat{\mathbf{b}}$ according to Eq. (\ref{eq what bhat solution}).
   \STATE Update $\hat{\mathbf{P}}$ according to Eq. (\ref{eq Q solution}).
   \STATE Update $\mathbf{G}$ according to Eq. (\ref{eq G solution}).
   \STATE Update $\mathbf{P}$ according to Eq. (\ref{eq P solution}).

   \UNTIL{Convergence}
   \STATE {\bfseries Return}: $y^\ast$ according to Eq. (\ref{eq prediction}).
\end{algorithmic}
\end{algorithm}

\section{Numerical Solution of the Partner Classifier in PLCP}
Th optimization problem in Eq. (\ref{eq: partner classifier}
) in the main paper has three variables with different constraints. We can adopt an alternative and iterative optimization to solve it. 

\subsection{Update $\hat{\mathbf{W}}$ and $\hat{\mathbf{b}}$}
With $\mathbf{C}$ fixed, the problem w.r.t. $\hat{\mathbf{W}}$ and $\hat{\mathbf{b}}$ can be written as

\begin{equation}
\min_{\hat{\mathbf{W}},\hat{\mathbf{b}}}\quad \left\|\mathbf{X}\hat{\mathbf{W}}+\mathbf{1}_n\hat{\mathbf{b}}^{\mathsf{T}}-\mathbf{C}\right\|_F^2 + \phi
\left\|\hat{\mathbf{W}}\right\|_F^2,
\label{eq: W subproblem}
\end{equation}
which is a least square problem with the closed-form solution as
\begin{equation}
\begin{split}
    \hat{\mathbf{W}} & = \left(\mathbf{X}^{\mathsf{T}}\mathbf{X} + \phi\mathbf{I}_{n\times n}\right )\mathbf{X}^{\mathsf{T}}\mathbf{C} \\ 
    \hat{\mathbf{b}} &= \frac{1}{n} \left(\mathbf{C}^{\mathsf{T}} \mathbf{1}_n - \hat{\mathbf{W}}^{\mathsf{T}}\mathbf{X}^{\mathsf{T}}\mathbf{1}_n\right),
\end{split} 
\label{eq w solution}
\end{equation}
where $\mathbf{I}_{n\times n}$ is the identity matrix with the size $n\times n$.

\subsection{Update $\mathbf{C}$}
With $\hat{\mathbf{W}}$ and $\hat{\mathbf{b}}$ fixed, the $\mathbf{C}$-subproblem can be formulated as 

\begin{equation}
\begin{split}
\min_{\mathbf{C}}&\quad \left\|\mathbf{X}\hat{\mathbf{W}}+\mathbf{1}_n\hat{\mathbf{b}}^{\mathsf{T}}-\mathbf{C}\right\|_F^2  + \gamma{\rm tr}\left(\mathbf{O}_1 \mathbf{C}^{\mathsf{T}}\right)\\
{\rm s.t.} & \quad \hat{\mathbf{Y}} \leq \mathbf{C} \leq \mathbf{1}_{n\times l}, \mathbf{C}\mathbf{1}_l = (l-1)\mathbf{1}_n.
\end{split}
\label{eq C subproblem}
\end{equation}
For simplicity, $\mathbf{O}_1$ is written as $\mathbf{O}$ and $\mathbf{J} = \mathbf{X}\hat{\mathbf{W}}+\mathbf{1}_n\hat{\mathbf{b}}^{\mathsf{T}}$. Notice that each row of $\mathbf{C}$ is independent to other rows, therefore the problem in Eq. (\ref{eq C subproblem}) can be solved row by row:

\begin{equation}
\begin{split}
\min_{\mathbf{C}_i}&\quad \mathbf{C}_i^{\mathsf{T}}\mathbf{C}_i + \left(\gamma\mathbf{O}_i-2\mathbf{J}_i \right)^{\mathsf{T}}\mathbf{C}_i\\
{\rm s.t.} & \quad \hat{\mathbf{Y}}_i \leq \mathbf{C}_i \leq \mathbf{1}_{n}, \mathbf{C}_i\mathbf{1}_l = l-1.
\end{split}
\label{eq C subproblem i}
\end{equation}
The problem in Eq. (\ref{eq C subproblem i}) is a standard Quadratic Programming (QP) problem, which can be solved by off-the-shelf QP tools. 

Note that this alternative and iterative optimization method is widely used in existing methods \cite{jia2023complementary,2018Leveraginglalo,feng2019partialsssfagdsg}.

\section{Extensions of PLCP}
\subsection{Kernel Extension \label{section kernel extension}}

Following a similar procedure in Sec. \ref{sec kernel}, we have

\begin{equation}
\begin{split}
    &\hat{\mathbf{W}} = \frac{\mathbf{\Phi}^{\mathsf{T}}\mathbf{A}}{2\lambda}, \mathbf{A} = \left(\frac{1}{2\lambda} \mathbf{K} + \frac{1}{2}\mathbf{I}_{n\times n}\right)^{-1}\left(\mathbf{C} - \mathbf{1}_n\hat{\mathbf{b}}^{\mathsf{T}}\right),\\
    &\mathbf{s} =\mathbf{1}_n^{\mathsf{T}}\left(\frac{1}{2\lambda} \mathbf{K} + \frac{1}{2}\mathbf{I}_{n\times n}\right)^{-1}, \hat{\mathbf{b}}^{\mathsf{T}} = \frac{\mathbf{s}\mathbf{C}}{\mathbf{s1}_n},
\end{split}
    \label{eq KKT solution}
\end{equation}
The modeling output is denoted by 

\begin{equation}
\hat{\mathbf{H}} =  \frac{1}{2\lambda}\mathbf{KA} + \mathbf{1}_n\hat{\mathbf{b}}^{\mathsf{T}}.
\label{eq Kernel modeling output}
\end{equation}

%and for an unseen sample $\mathbf{x}^{\ast}$, the predicted label $y_{p}$ is
%\begin{equation}
%    y_{p} = {\rm argmin}_k \sum^n_{i=1}a_{ik}\mathcal{K}(\mathbf{x}^{\ast}, \mathbf{x}_i)+\hat{b}_k.
%\end{equation}

\subsection{Deep-learning Extension \label{section deep-learning extension}}
%To further demonstrate the universality of our proposed learning paradigm, we can extent PLCP to deep-learning based approaches by adding an absolutely same model that contains both a feature encoder and a classification head like those in original methods as the partner classifier. Another modification is the formulation of the loss function on which we append two extra losses.

\subsubsection{Extension to Deep-learning Based Method \label{section deep-learning extension detail}}
PLCP can be also extended to a deep-learning version to further demonstrate its universality and effectiveness. The model of the majority of the deep-learning based methods usually contains a feature encoder followed by a prediction head, which predicts the labeling confidence of each sample. Denote a model in $\mathcal{B}$ with such architecture $g(\cdot)$, specifically, an additional model $\hat{g}(\cdot)$ with the same architecture as $g(\cdot)$ is introduced as the partner classifier, which predicts the non-candidate labeling confidence of each sample. Given an example $\mathbf{x}$ and its candidate label vector $\mathbf{y}$, the non-candidate loss is defined as
\begin{equation}
     \mathcal{L}_{com} =  - \sum_{i=1}^l (1-y_i) \log(\hat{g}_i(\mathbf{x})).
    \label{eq cross entropy }
\end{equation}
Afterwards, a collaborative loss is designed to link the two models:
\begin{equation}
\mathcal{L}_{col} = \sum_{i=1}^l p_i\hat{p}_i.
    \label{eq colla loss}
\end{equation}
Here, $\mathbf{p} = [p_1,p_2,...,p_l]$ and $\hat{\mathbf{p}} =[\hat{p}_1,\hat{p}_2,...,\hat{p}_l]$ are two blurred predictions of $g(\mathbf{x})$ and $\hat{g}(\mathbf{x})$ respectively, where
\begin{equation}
\begin{split}
    \mathbf{p} &= \frac{\phi(e^k g(\mathbf{x})) \odot \mathbf{y}}{(\phi(e^k g(\mathbf{x})) \odot \mathbf{y})\mathbf{1}_l} \\
    \hat{\mathbf{p}} & = \mathbf{1}_l^{\mathsf{T}} - \frac{\phi(e^k (1-\hat{g}(\mathbf{x}))) \odot \mathbf{y}}{(\phi(e^k (1-\hat{g}(\mathbf{x}))) \odot \mathbf{y})\mathbf{1}_l}.
\end{split}
    \label{eq blur loss}
\end{equation}

The overall loss function is:
\begin{equation}
    \mathcal{L} = \mathcal{L}_{ori} + \mathcal{L}_{com} + \mu \mathcal{L}_{col},
    \label{eq loss function}
\end{equation}
where $\mathcal{L}_{ori}$ is the original loss function of $\mathcal{B}$ and $\mu$ is a hyper-parameter.

\subsubsection{Implementation Details \label{section imple}}
In PLCP, we set $k=-1$ and $\mu=0.5$. All the hyper-parameters and experimental setups of $\mathcal{B}$ are set based on their original papers.

\bibliographystyle{IEEEtran}
\bibliography{main}